\documentclass[pdflatex,sn-mathphys]{sn-jnl}

\jyear{2022}%

\usepackage{amsmath}
\usepackage{multirow}
\usepackage{enumitem}

\DeclareMathOperator*{\argmin}{argmin}
\DeclareMathOperator*{\argmax}{argmax}
\DeclareMathOperator*{\vct}{vec}
\DeclareMathOperator*{\sgn}{sgn}
\DeclareMathOperator*{\tr}{tr}
\DeclareMathOperator*{\cov}{cov}

\newcommand{\ind}{\perp\!\!\!\!\perp}

\newcommand{\cD}{\mathcal{D}}
\newcommand{\cX}{\mathcal{X}}
\newcommand{\cY}{\mathcal{Y}}

\newcommand{\cU}{\mathcal{U}}
\newcommand{\cV}{\mathcal{V}}

\newcommand{\cM}{\mathcal{M}}
\newcommand{\cB}{\mathcal{B}}
\newcommand{\cO}{\mathcal{O}}
\newcommand{\cS}{\mathcal{S}}

\newcommand{\bx}{\mathbf{x}}
\newcommand{\bE}{\mathbb{E}}

\newcommand{\bR}{\mathbb{R}}

\newcommand{\eps}[2]{\boldsymbol{\epsilon}^{(#1)}_{(#2)^*}}
\newcommand{\xc}[2]{\mathbf{x}^{(#1)}_{(#2)^*}}
\newcommand{\my}[2]{m^{#1}(y^{(#1)}_{(#2)})}

\newcommand{\ep}[3]{\boldsymbol{\epsilon}^{(#1)}_{(#2)^*,#3}}
\newcommand{\m}[3]{m^{#1}_{#3}(y^{(#1)}_{(#2)})}

\theoremstyle{thmstyleone}%
\newtheorem{theorem}{Theorem}
\newtheorem{lemma}[theorem]{Lemma}
\newtheorem{corollary}[theorem]{Corollary}
\newtheorem{proposition}[theorem]{Proposition}%

\theoremstyle{thmstyletwo}%
\newtheorem{remark}{Remark}%

\theoremstyle{thmstylethree}%
\newtheorem{definition}{Definition}%

\begin{document}

\title[Federated Sparse Sliced Inverse Regression]{Federated Sufficient Dimension Reduction Through High-Dimensional Sparse Sliced Inverse Regression}


\author*[1]{\fnm{Wenquan} \sur{Cui}}\email{wqcui@ustc.edu.cn}

\author[1]{\fnm{Yue} \sur{Zhao}}\email{zhy99@mail.ustc.edu.cn}

\author[1]{\fnm{Jianjun} \sur{Xu}}\email{xjj1994@mail.ustc.edu.cn}

\author[2]{\fnm{Haoyang} \sur{Cheng}}\email{chyling@mail.ustc.edu.cn}

\affil*[1]{\orgdiv{International Institute of Finance, School of Management}, \orgname{University of Science and Technology of China}, \orgaddress{\city{Hefei}, \postcode{230026}, \state{Anhui}, \country{People's Republic of China}}}

\affil[2]{\orgdiv{College of Electrical and Information Engineering}, \orgname{Quzhou University}, \orgaddress{\city{Quzhou}, \postcode{324000}, \state{Zhejiang}, \country{People's Republic of China}}}


\abstract{Federated learning has become a popular tool in the big data era nowadays. 
It trains a centralized model based on data from different clients while keeping data decentralized. 
In this paper, we propose a federated sparse sliced inverse regression algorithm for the first time. 
Our method can simultaneously estimate the central dimension reduction subspace and perform variable selection in a federated setting. 
We transform this federated high-dimensional sparse sliced inverse regression problem into a convex optimization problem by constructing the covariance matrix safely and losslessly. 
We then use a linearized alternating direction method of multipliers algorithm to estimate the central subspace. 
We also give approaches of Bayesian information criterion and hold-out validation to ascertain the dimension of the central subspace and the hyper-parameter of the algorithm. 
We establish an upper bound of the statistical error rate of our estimator under the heterogeneous setting. 
We demonstrate the effectiveness of our method through simulations and real world applications. }




\maketitle

\section{Introduction}\label{sec1}

Federated learning is a distributed machine learning paradigm that collaboratively trains a model with data on many clients. 
Unlike traditional distributed machine learning methods, which partition data into different clients to improve the efficiency of the learning algorithm, the goal of federated learning is to solve the learning problem without requiring the clients to reveal too much local information. 
With the increasing demand for data security and privacy protection, federated learning has received significant attention in both industry and academia. 
For example, banks want to collaboratively train a credit card scoring model without disclosing information about their customers, or hospitals want to carry out researches on a rare disease with each other due to the small number of sample cases, but they can't expose their patients' identity. 
For more on the progress of federated learning, see \cite{Kairouz2019, Yang2020FL}. 

The term \textit{federated learning} was introduced by McMahan et al. \cite{McMahan2016}, they also proposed the Federated Averaging (FedAvg) algorithm. 
FedAvg composes multiple rounds of local stochastic gradient descent updates and server-side averaging aggregation to train a centralized model. 
FedAvg and its variants such as FedProx \citep{LiTian2018FedProx}, SCAFFOLD \citep{Mohri2019SCAFFOLD} and FedAc \citep{yuan2020fedac} mainly focus on deep neural networks, while federated adaptations of traditional machine learning methods are rarely studied. 
Especially in the area of dimension reduction and variable selection, few researches have appeared under federated learning settings. 
Grammenos et al. \cite{grammenos2020fpca} proposed the federated principal component analysis (FedPCA) to reduce the dimensionality of the data. 
Chai et al. \cite{chai2021fedsvd} offered masking-based federated singular vector decomposition (FedSVD) method, which can also perform dimension reduction by picking the $k$ right singular vectors with the largest singular values as the projection matrix. 
However, principal component analysis is an unsupervised learning method that doesn't take into account the relationship between the responses and covariates. 
Federated learning mainly applies to regression or classification problems with response variables. 
Thus, we would like to develop a federated sufficient dimension reduction method that can reflect the relationship between responses and covariates. 

Sufficient dimension reduction \citep{Li1991SIR, Cook1994plot} aimed at reducing the dimension of data without loss of sufficient information. 
Consider a univariate response $y \in \bR$ combined with a stochastic covariate vector $\bx = (x_1, \dots, x_d)^{\mathrm{T}} \in \bR^d$. 
Let $K < d$ and $\mathbf{B} = (\boldsymbol{\beta}_1, \dots, \boldsymbol{\beta}_K) \in \bR^{d \times K}$ such that
\begin{equation}
    \label{eq2.1}
    y \ind \bx \mid (\mathbf{B}^{\mathrm{T}} \bx), 
\end{equation}
where $\ind$ signifies statistical independence. 
Equation (\ref{eq2.1}) implies that $y \vert \bx$ and $y \vert (\mathbf{B}^{\mathrm{T}} \bx)$ have identical distribution. 
In other words, $\mathbf{B}^{\mathrm{T}} \bx$ summarizes information in $\bx$ with respect to $y$. 
Therefore, it is sufficient to replace $\bx$ with a set of $K$ linear combinations of $\bx$ to characterize the dependence of $y$ on $\bx$. 
$\boldsymbol{\beta}_1, \dots, \boldsymbol{\beta}_K$ are defined as sufficient dimension reduction directions. 
In general, the dimension reduction space spanned by $\mathbf{B}$ is not unique. 
Define the intersection of all dimension reduction subspaces as the central subspace $\cV_{y \vert \bx}$ \citep{Cook1994plot}. 
Originally from \cite{Li1991SIR}, a general regression model was proposed to characterize the relationship between $y$ and $\bx$: 
\begin{equation}
    \label{eq2.2}
    y = f(\mathbf{B}^{\mathrm{T}} \bx, \varepsilon), 
\end{equation}
where $f$ is an unknown link function and $\varepsilon$ captures stochastic noises. 
Zeng and Zhu \cite{Zeng2010integral} proved that model (\ref{eq2.2}) is equivalent to model (\ref{eq2.1}) in the sense that a set of $K$ linear combinations of $\bx$ captures the conditional distribution of $y$ given $\bx$. 
We usually consider the estimation of $\cV_{y \vert \bx}$. 
Many estimation approaches of the central subspace have been proposed in the literature, including sliced inverse regression (SIR) \citep{Li1991SIR}, sliced average variance estimates (SAVE) \citep{Cook2000SAVE}, inverse regression (IR) \citep{Cook2005IR}, direction regression (DR) \citep{Li2007DR}, principal Hessian directions (PHD) \citep{Li1992PHD}, minimum average variance estimation (MAVE) \citep{xia2002MAVE}, etc. 

SIR and its variants are perhaps the most widely used among all these methods and we will focus on SIR in this paper. 
In \cite{hsing1992asym, zhu1995asymptotics, zhu2006sirhdc}, authors have studied methods and asymptotics for SIR. 
However, SIR estimations usually contain all $d$ covariates, making the dimension reduction hard to interpret. 
Researchers have made some attempts to perform variable selection via sparse sliced inverse regression, see \cite{li2005mf, li2008rsir, Lin2019LassoSIR}. 
These methods conduct sparse dimension reduction in a step-wise process, which means they estimate a sparse solution for each direction one at a time. 
Chen et al. \cite{chen2010coordinate} added a penalty term to encourage sparsity while estimating central subspace directly. 
Tan et al. \cite{Kean2018convex} turned it into a convex optimization problem and used the linearized alternating direction method of multipliers (ADMM) algorithm \citep{Boyd2011ADMM, Fang2015GADMM} to solve the optimization problem. 

Our proposal is the federated fashion of sparse sliced inverse regression in the high-dimensional setting. 
We rewrite sliced inverse regression in a weighted averaging form and use a modified version of the FedSVD algorithm to construct the covariance matrix losslessly and securely. 
Through convex relaxation of the constraint, we can turn the federated sliced inverse regression problem into a convex optimization problem. 
We add a $L_1$ regularization term to the optimization objective to yield sparse solutions. 
For this optimization problem, we use the linearized ADMM algorithm to solve it, which has been proved helpful in the federated learning literature, see \cite{zhang2020fedpd}. 
We give an approach of Bayesian information criterion (BIC) to determine the dimension of the central subspace under the federated setting. 
We use hold-out validation to select the tuning parameter of the ADMM algorithm. 
Our method only needs to transfer intermediate variables and masked data between server and clients and does not require communication of the raw data or other variables, ensuring privacy protection and data security. 
We also analyze the upper error bound of our estimator in the non-i.i.d. and high-dimensional setting. 
Our method applies to a broader range of conditions, allowing different distributions of responses and covariates across clients. 

The outline of this article is organized as follows. 
In Section \ref{scenario}, we introduce the problem setup of federated sparse sliced inverse regression and give our FedSSIR algorithm for estimating the central subspace. 
Section \ref{theoretical} provides an upper bound on the subspace distance between FedSSIR estimation and the true subspace. 
Numerical simulations and real data applications are presented in Section \ref{numerical}. 
Section \ref{conclusion} finishes this article with a brief conclusion and discussion on possible extensions. 

\section{Learning Scenario}
\label{scenario}
In this section, we describe the learning scenario of federated sparse sliced inverse regression. 

We start with some general definitions and notations used throughout the paper. 
Assume we have $m$ clients, and the $i$-th client holds dataset $S_i$, $\forall i \in [m]$, where $[n]$ denotes the set $\{1, \dots, n\}$. 
$S_i$ is a set of $n_i$ samples $(\bx^{(i)}_1, y^{(i)}_1), \dots, (\bx^{(i)}_{n_i}, y^{(i)}_{n_i}) \in \cX \times \cY$, which are i.i.d. drawn from distribution $\cD_i$. 
For input space $\cX \subseteq \bR^d$ and output space $\cY \subseteq \bR$, $\bx^{(i)} = (x_1^{(i)}, \dots, x_d^{(i)})^{\mathrm{T}} \in \cX$ is a stochastic covariate vector, $y^{(i)} \in \cY$ is a univariate response. 
$N = \sum_{i=1}^{m} n_i$ denotes the number of all samples. 
We denote the mean and covariance matrix of $\bx^{(i)}$ as $\boldsymbol{\mu}_i$ and $\boldsymbol{\Sigma}_i$, and the covariance matrix of the conditional expectation $\mathbf{T}_i = \cov(\bE[\bx^{(i)} \vert y^{(i)}])$. 
Define the $i$-th sample mean and sample covariance matrix as $\bar{\bx}^{(i)} = \frac{1}{n_i} \sum_{j=1}^{n_i} \bx^{(i)}_j$ and $\hat{\boldsymbol{\Sigma}}_i = \frac{1}{n_i} \sum_{j=1}^{n_i} (\bx^{(i)}_j - \bar{\bx}^{(i)}) (\bx^{(i)}_j - \bar{\bx}^{(i)})^{\mathrm{T}}$. 

For a vector $\mathbf{v}$, we use $\|\mathbf{v}\|_0, \|\mathbf{v}\|_1, \|\mathbf{v}\|_2, \|\mathbf{v}\|_{\infty}$ to denote the $L_0$ norm, $L_1$ norm, $L_2$ norm and $L_{\infty}$ norm of $\mathbf{v}$, respectively. 
For a matrix $\mathbf{M}$, we use $\|\mathbf{M}\|_*, \|\mathbf{M}\|_{\mathrm{F}}, \|\mathbf{M}\|_{1,1}, \|\mathbf{M}\|_2, \|\mathbf{M}\|_{\max}$ to denote the nuclear norm, Frobenius norm, $L_{1,1}$ norm, $L_2$ norm and max norm of $\mathbf{M}$, respectively, where $\|\mathbf{M}\|_* = \tr (\sqrt{\mathbf{M}^{\mathrm{T}} \mathbf{M}})$, $\|\mathbf{M}\|_{\mathrm{F}} = \sqrt{\tr (\mathbf{M}^{\mathrm{T}} \mathbf{M})}$, $\|\mathbf{M}\|_{1,1} = \sum_{i,j} \lvert \mathbf{M}_{i,j} \rvert$, $\|\mathbf{M}\|_2 = \max_{v: \|v\|_2 = 1} \|\mathbf{M} v\|_2$, $\|\mathbf{M}\|_{\max} = \max_{i,j} \lvert \mathbf{M}_{i,j} \rvert $. 
$\vct(\mathbf{M})$ is the vectorization operator which stacks the columns of $\mathbf{M}$ into a vector. 
For two matrices $\mathbf{A}$ and $\mathbf{B}$, we denote their inner product as $\langle \mathbf{A}, \mathbf{B} \rangle = \sqrt{\tr(\mathbf{A}^{\mathrm{T}} \mathbf{B})}$. 
$\mathbf{A} \otimes \mathbf{B}$ denotes the Kronecker product of matrices $\mathbf{A}$ and $\mathbf{B}$. 

\subsection{Problem Formulation}
\label{sec2.1}

We first introduce the formulation of the federated sliced inverse regression problem. 
We can regard federated learning as a learning paradigm on a mixture distribution $\cD$ consisting of client specific distributions, which is $\cD = \sum_{i=1}^{m} \omega_i \cD_i$, where $\omega_i$ is the mixture weight of $\cD_i$ and $\sum_{i=1}^{m} \omega_i = 1$. 
In order to represent this mathematically, we introduce a latent variable $z$ as an indicator of client. 
We can represent the distribution as
\begin{align*}
    \mathrm{Pr}(z = i) = \omega_i,\quad & \forall i \in [m], \\
    \bx, y \vert z = i \sim \cD_i,\quad & \forall i \in [m]. 
\end{align*}
Here, we give formulations of the covariance matrix and the covariance matrix of the conditional expectation under this mixture distribution. 
Assuming that the mean of the covariate satisfies $\boldsymbol{\mu}_1 = \cdots = \boldsymbol{\mu}_m = 0$, in practical, we only need to centralize the data on each client. 
The population covariance matrix can be formulated as 
\begin{equation}
\begin{aligned}
    \mathrm{cov}(\bx) & = \bE[\mathrm{cov}(\bx \vert z)] + \mathrm{cov}(\bE[\bx \vert z]) \\
    & = \sum_{i=1}^{m} \omega_i \boldsymbol{\Sigma}_i + \sum_{i=1}^{m} \omega_i (\boldsymbol{\mu}_i - \sum_{i=1}^{m} \omega_i \boldsymbol{\mu}_i)^2 \\
    & = \sum_{i=1}^{m} \omega_i \boldsymbol{\Sigma}_i. 
\end{aligned}
\end{equation}
Let $m(y) = \bE[\bx \vert y]$, the population covariance matrix of the conditional expectation can be formulated as
\begin{equation}
\begin{aligned}
    \mathrm{cov}(m(y)) & = \bE[\mathrm{cov}(m(y) \vert z)] + \mathrm{cov}(\bE[m(y) \vert z]) \\
    & = \sum_{i=1}^{m} \omega_i \mathbf{T}_i + \sum_{i=1}^{m} \omega_i (\boldsymbol{\mu}_i - \sum_{i=1}^{m} \omega_i \boldsymbol{\mu}_i)^2 \\
    & = \sum_{i=1}^{m} \omega_i \mathbf{T}_i, 
\end{aligned}
\end{equation}
where the second equation comes from $\bE[\bE[m(y) \vert z]] = \bE[m(y)] = \bE[\bE[\bx \vert y]] = \bE[\bx]$ and $\bE[m(y)\vert z = i] = \bE[\bE[\bx^{(i)} \vert y^{(i)}]] = \bE[\bx^{(i)}]$. 
We denote $\mathrm{cov}(m(y))$ as $\bar{\mathbf{T}}$. 
Therefore, the empirical covariance matrix $\hat{\boldsymbol{\Sigma}} = \sum_{i=1}^{m} \hat{\omega}_i \hat{\boldsymbol{\Sigma}}_i$ and the empirical covariance matrix of the conditional expectation $\hat{\mathbf{T}} = \sum_{i=1}^{m} \hat{\omega}_i \hat{\mathbf{T}}_i$. 
Under this representation, we have $n_i = \sum_{j=1}^{N} 1_{(z_j = i)}$, $\forall i \in [m]$. 
$(n_1, \dots, n_m)$ has a multinomial distribution $\mathrm{Multinomial}(N, \omega_1, \dots, \omega_m)$, which means $\omega_i$ can be estimated by $n_i / N$. 
Thus, we have $\hat{\boldsymbol{\Sigma}} = \sum_{i=1}^{m} \frac{n_i}{N} \hat{\boldsymbol{\Sigma}}_i$ and $\hat{\mathbf{T}} = \sum_{i=1}^{m} \frac{n_i}{N} \hat{\mathbf{T}}_i$. 

Suppose distribution $\cD_i$ satisfies model (\ref{eq2.2}) with client specific link function $f_i$ and noise $\varepsilon_i$: 
\begin{equation}
    \label{eq2.3}
    y^{(i)} = f_i(\mathbf{B}^{\mathrm{T}} \bx^{(i)}, \varepsilon_i), \forall i \in [m].
\end{equation}
Model (\ref{eq2.3}) implies that under certain conditions, data distributions on different clients may vary from each other in our federated learning system, as long as all clients share the same central subspace. 
For example, the feature vectors of the user groups of two banks share the same internal structure, although their label dependencies on the central subspace have personal or regional variation. 
This feature of our method can help us deal with non-i.i.d. data, especially for the skewed conditional distribution of $y^{(i)}$ given $\bx^{(i)}$, which is known as \textit{concept shift} in federated learning literature. 
For further discussion on non-i.i.d. data in federated learning, see \cite{hsieh2020noniid, Kairouz2019}. 

As for SIR, it requires the covariate $\bx^{(i)}$ to satisfy the linear condition: 
\begin{equation}
    \label{eq2.4}
    \bE[\bx^{(i)} \vert \mathbf{B}^{\mathrm{T}} \bx^{(i)}] = \mathbf{b}^{(i)} + \mathbf{W}^{(i)} \mathbf{B}^{\mathrm{T}} \bx^{(i)}, \forall i \in [m], 
\end{equation}
where $\mathbf{b}^{(i)} \in \bR^{d}$ and $\mathbf{W}^{(i)} \in \bR^{d \times K}$ are some constants. 
According to Theorem 3.1 and Corollary 3.1 in \citep{Li1991SIR}, under model (\ref{eq2.3}) and condition (\ref{eq2.4}), assuming that each $\cD_i$ has the same covariance matrix $\boldsymbol{\Sigma}$, the inverse regression curve $\bE[\bx^{(i)} \vert y^{(i)}] - \bE[\bx^{(i)}]$ is contained in the column space of $\boldsymbol{\Sigma} \mathbf{B}$, which means the covariance matrix of the conditional expectation $\mathbf{T}_i$ degenerates in the orthogonal directions of $\boldsymbol{\Sigma} \boldsymbol{\beta}_k$'s. 
Thus, recall that $\bar{\mathbf{T}} = \sum_{i=1}^{m} \omega_i \mathbf{T}_i$, $\bar{\mathbf{T}}$ also degenerates in the orthogonal directions of $\boldsymbol{\Sigma} \boldsymbol{\beta}_k$'s. 

The linear condition is satisfied when the covariate $\bx^{(i)}$ is elliptically distributed \citep{EATON1986ellipical}. 
The linear condition may be considered to be strict for real data, whose distribution is elusive for us to catch. 
Hall and Li \cite{hall1993almost} argued that the condition is approximately satisfied when the dimension of data is large, regardless of the actual distribution of data. 
From the discussion above, we can see that our method does not need the assumption that all clients have the identical marginal distribution for $\bx$, provided each $\cD_i$ satisfies condition (\ref{eq2.4}) and has the same covariance matrix. 
This kind of non-i.i.d.-ness is known as \textit{covariate shift} in federated learning. 
Covariate shift, according to Hidetoshi~\cite{Hidetoshi2000covariate}, means that covariate on each client has different marginal distributions. 
For example, patients in hospitals in two different countries may be biased in physical conditions. 
Numerical experiments of our method against non-i.i.d. data have been carried out in Section \ref{numerical}. 

\subsection{Federated Sparse Sliced Inverse Regression}
\label{sec2.2}

Then we consider the sliced inverse regression problem under the non-i.i.d. setting. 
Assume that for each client $i$, $\cD_i$ satisfies model (\ref{eq2.3}) and condition (\ref{eq2.4}), and has the same covariance matrix $\boldsymbol{\Sigma}$. 
Then we have $\bar{\mathbf{T}} \boldsymbol{\beta}_k = \lambda_k \boldsymbol{\Sigma} \boldsymbol{\beta}_k$ for $k \in [K]$, where $\boldsymbol{\beta}_k^{\mathrm{T}} \boldsymbol{\Sigma} \boldsymbol{\beta}_k = 1$, $\boldsymbol{\beta}_k^{\mathrm{T}} \boldsymbol{\Sigma} \boldsymbol{\beta}_l = 0$ for $l \neq k$, $\lambda_k$ is the k-th largest generalized eigenvalue. 

In the traditional setting, we have access to all data, and we can get the estimators $\hat{\boldsymbol{\Sigma}}$ and $\hat{\mathbf{T}}$ based on the full sample. 
In the federated setting, we can't aggregate all data in one server, and client $i$ can only access its own $\hat{\boldsymbol{\Sigma}}_i$ and $\hat{\mathbf{T}}_i$. 
Recall that $\hat{\mathbf{T}} = \frac{1}{N} \sum_{i=1}^{m} n_i \hat{\mathbf{T}}_i$, $\hat{\boldsymbol{\Sigma}} = \frac{1}{N} \sum_{i=1}^{m} n_i \hat{\boldsymbol{\Sigma}}_i$. 
Then a basis $\mathbf{V}$ of the central subspace can be estimated by solving a weighted averaging version of the generalized eigenvalue problem
\begin{equation}
    \label{eq2.5c}
    \hat{\mathbf{T}} \mathbf{V} = \hat{\boldsymbol{\Sigma}} \mathbf{V} \mathbf{\Lambda}, 
\end{equation}
where $\mathbf{V}^{\mathrm{T}} \hat{\boldsymbol{\Sigma}} \mathbf{V} = \mathbf{I}_K$, $\mathbf{\Lambda}$ is a diagonal matrix with elements $\{\lambda_1, \dots, \lambda_K\}$. 

The generalized eigenvalue problem (\ref{eq2.5c}) can be equivalently converted into a non-convex optimization problem
\begin{equation}
    \label{eq2.7}
    \min_{\mathbf{V} \in \bR^{d \times K}} - \tr \{ \mathbf{V}^{\mathrm{T}} \hat{\mathbf{T}} \mathbf{V} \} \text{ subject to } \mathbf{V}^{\mathrm{T}} \hat{\boldsymbol{\Sigma}} \mathbf{V} = \mathbf{I}_K. 
\end{equation}
For privacy preserving, we prefer not to share local covariance matrices $\hat{\boldsymbol{\Sigma}}_i$'s and covariance matrices of the conditional expectation $\hat{\mathbf{T}}_i$'s between clients and server either. 

Let $\boldsymbol{\Pi} = \mathbf{V} \mathbf{V}^{\mathrm{T}}$ and use the property of trace operator, (\ref{eq2.7}) turns into
\begin{equation}
    \label{eq2.8}
    \min_{\boldsymbol{\Pi} \in \cB} - \sum_{i=1}^{m} \frac{n_i}{N} \tr \{ \hat{\mathbf{T}}_i \boldsymbol{\Pi} \}, 
\end{equation}
where $\cB = \{\boldsymbol{\Pi} = \mathbf{V} \mathbf{V}^{\mathrm{T}}: \mathbf{V}^{\mathrm{T}} \hat{\boldsymbol{\Sigma}} \mathbf{V} = \mathbf{I}_K \}$. 
(\ref{eq2.8}) consists of additive objectives $\sum_{i=1}^{m} \tr \{ \hat{\mathbf{T}}_i \boldsymbol{\Pi} \}$. 
In the update step of our algorithm, client $i$ only needs to handle the local objective involving $\hat{\mathbf{T}}_i$, and the server will aggregate the intermediate results. 
This means that our method does not need to compute $\hat{\mathbf{T}}$, but only $\hat{\mathbf{T}}_i$ on each client. 

How to construct the covariance matrix $\hat{\boldsymbol{\Sigma}}$ is another problem we have to face. 
If we deal with the covariance matrix separately on each client as we did with the covariance matrix of the conditional expectation, the constraints can be very complicated. 
Thus, we use a modified version of FedSVD \citep{chai2021fedsvd} to securely and losslessly construct $\hat{\boldsymbol{\Sigma}}$, see Algorithm \ref{alg0}. 
The proof of lossless precision for this masking based method can be found in Theorem 4.1 in \cite{chai2021fedsvd}. 

\begin{algorithm}[ht]
\caption{Covariance estimation via FedSVD}
\label{alg0}
\begin{algorithmic}[1]
\Require centralized covariates $\mathbf{X}_i = (\bx^{(i)}_1, \dots, \bx^{(i)}_{n_i}) - \bar{\bx}^{(i)} \mathbf{1}^{\mathrm{T}}$, $i \in [m]$
\Ensure the average covariance matrix $\hat{\boldsymbol{\Sigma}}$
\State \textbf{Server do: } 
    \State Generate random orthogonal matrix $\mathbf{P}$; 
    \State Broadcast $\mathbf{P}$ to clients; 
\For{\textbf{Client} $i \in [m]$ in parallel}
    \State Generate random orthogonal matrix $\boldsymbol{\Psi}_i$; 
    \State Compute $\mathbf{X}'_i = \mathbf{P} \mathbf{X}_i \boldsymbol{\Psi}_i$; 
    \State Send $\mathbf{X}'_i$ to Server; 
\EndFor
\State \textbf{Server do: }
    \State Aggregate $\mathbf{X}' = (\mathbf{X}'_1, \dots, \mathbf{X}'_m)$; 
    \State Factorize $\mathbf{X}'$ into $[\mathbf{U}', \mathbf{D}, \sim]$ via SVD; 
    \State Recover $\mathbf{U}$ through $\mathbf{P}^{\mathrm{T}} \mathbf{U}'$; 
    \State Compute $\hat{\boldsymbol{\Sigma}}$ through $\frac{1}{N} \mathbf{U} \mathbf{D}^2 \mathbf{U}^{\mathrm{T}}$; 
\end{algorithmic}
\end{algorithm}

Similar to \cite{vu2013fantope}, we propose a convex relaxation for the non-convex optimization problem in (\ref{eq2.8})
\begin{equation}
    \label{eq2.9}
    \begin{aligned}
        & \min_{\boldsymbol{\Pi} \in \cM} - \sum_{i=1}^{m} \frac{n_i}{N} \tr \{ \hat{\mathbf{T}}_i \boldsymbol{\Pi} \} \\ 
        & \text{ subject to } \|\hat{\boldsymbol{\Sigma}}^{1/2} \boldsymbol{\Pi} \hat{\boldsymbol{\Sigma}}^{1/2} \|_* \leq K, \|\hat{\boldsymbol{\Sigma}}^{1/2} \boldsymbol{\Pi} \hat{\boldsymbol{\Sigma}}^{1/2} \|_2 \leq 1, \\
    \end{aligned}
\end{equation}
where $\cM$ is the set of all $d \times d$ positive semi-definite matrices. 

\begin{remark}
Under (\ref{eq2.8}), $\mathbf{V}$ belongs to the set $\{\mathbf{V} \in \bR^{d \times K}: \mathbf{V}^{\mathrm{T}} \hat{\boldsymbol{\Sigma}} \mathbf{V} = \mathbf{I}_K \}$, which is an embedded submanifold of $\bR^{d \times K}$. 
There are some algorithms proposed to solve optimization problems on matrix manifolds, see \cite{AbsilMahonySepulchre2009manifolds}. 
However, we choose to adopt the convex relaxation like (\ref{eq2.9}) in this paper because of the advantages of convex optimization. 
\end{remark}

\begin{remark}
Traditional sufficient dimension reduction methods usually need to compute the inverse of covariance matrix \citep{Li1991SIR, Li1992PHD, Cook2000SAVE, Li2007DR}, while (\ref{eq2.9}) doesn't involve the inversion of $\hat{\boldsymbol{\Sigma}}$. 
This can be helpful when the sample size $n$ is less than the dimension $d$. 
\end{remark}

In addition to dimension reduction, variable selection is an important goal of statistical analysis. 
To encourage sparsity, we can add a $L_1$ penalty on all elements of $\boldsymbol{\Pi}$
\begin{equation}
    \label{eq2.10}
    \begin{aligned}
        & \min_{\boldsymbol{\Pi} \in \cM} \sum_{i=1}^{m} \frac{n_i}{N} (- \tr \{ \hat{\mathbf{T}}_i \boldsymbol{\Pi} \} + \rho \| \boldsymbol{\Pi} \|_{1,1}) \\
        & \text{ subject to } \|\hat{\boldsymbol{\Sigma}}^{1/2} \boldsymbol{\Pi} \hat{\boldsymbol{\Sigma}}^{1/2} \|_* \leq K, \|\hat{\boldsymbol{\Sigma}}^{1/2} \boldsymbol{\Pi} \hat{\boldsymbol{\Sigma}}^{1/2} \|_2 \leq 1, \\
    \end{aligned}
\end{equation}
where $\rho$ is a positive tuning parameter. 
Larger $\rho$ will yield a sparser estimator of basis vectors. 
In practice, we can allow different $\rho_i$ to control the sparsity of intermediate results on client $i$ during algorithm iteration. 

\subsection{Federated Sparse Sliced Inverse Regression algorithm}

\begin{algorithm}[ht]
\caption{Federated Sparse Sliced Inverse Regression (FedSSIR)}
\label{alg1}
\begin{algorithmic}[1]
\Require $S_i = \{(\bx^{(i)}_j, y^{(i)}_j) \}_{j=1}^{n_i}$, $i \in [m]$, number of sufficient dimension reduction directions $K$, the tuning parameter $\rho$, the ADMM parameter $\nu > 0$, tolerance level $\varepsilon > 0$. 
\Ensure estimation of projection matrix onto the central subspace $\hat{\boldsymbol{\Pi}}$. 
\State Compute $\hat{\boldsymbol{\Sigma}}_i$, $\hat{\mathbf{T}}_i$ and the linearization parameter $\alpha_i = 4 \nu \lambda^2_{\max}(\hat{\boldsymbol{\Sigma}}_i)$ on each client, where $\lambda_{\max}(\mathbf{M})$ is the largest eigenvalue of $\mathbf{M}$; 
\State Use \texttt{FedSVD} to compute $\hat{\boldsymbol{\Sigma}}$; 
\State \textbf{Server} initialize the parameters: primal variables $\boldsymbol{\Pi}^0 = \mathbf{I}_d$, $H^0 = \mathbf{I}_d$, dual variable $\boldsymbol{\Gamma}^0 = 0$, intermediate variable $\mathbf{M}^0 = \hat{\boldsymbol{\Sigma}}^{2} - \hat{\boldsymbol{\Sigma}}$ and broadcast $\boldsymbol{\Pi}^0$, $\mathbf{M}^0$ to clients; 
\While{$\|\boldsymbol{\Pi}^t - \boldsymbol{\Pi}^{t-1} \|_{\mathrm{F}} > \varepsilon $}
    \For{\textbf{Client} $i \in [m]$ in parallel}
        \State $\boldsymbol{\Pi}_i^{t+} = \mathrm{ST}[\boldsymbol{\Pi}^t + \frac{1}{\alpha_i} \hat{\mathbf{T}}_i - \frac{\nu}{\alpha_i} \mathbf{M}^{t}, \frac{\rho}{\alpha_i} ]$, where $\mathrm{ST}$ stands for soft-thresholding operator element-wise applied to a matrix; 
        \State Send $\boldsymbol{\Pi}_i^{t+}$ to server; 
    \EndFor
    \State \textbf{Server updates: }
    \State $\boldsymbol{\Pi}^{t+1} = \sum_{i=1}^{m} \frac{n_i}{N} \boldsymbol{\Pi}_i^{t+}; $
    \State $\mathbf{H}^{t+1} = \sum_{i=1}^{d} \min \{1, \max(w_j - \gamma^*, 0)\} \mathbf{u}_j \mathbf{u}_j^{\mathrm{T}}$, where $\sum_{j=1}^d w_j \mathbf{u}_j \mathbf{u}_j^{\mathrm{T}}$ is the singular value decomposition of $\hat{\boldsymbol{\Sigma}}^{1/2} \boldsymbol{\Pi}^{t+1} \hat{\boldsymbol{\Sigma}}^{1/2} - \boldsymbol{\Gamma}^t$, and $$\gamma^* = \argmin_{\gamma > 0} \gamma, \quad \text{ subject to} \sum_{j=1}^d \min \{1, \max(w_j - \gamma, 0)\} \leq K; $$
    \State $\boldsymbol{\Gamma}^{t+1} = \boldsymbol{\Gamma}^t + \hat{\boldsymbol{\Sigma}}^{1/2} \boldsymbol{\Pi}^{t+1} \hat{\boldsymbol{\Sigma}}^{1/2} - \mathbf{H}^{t+1} $; 
    \State $\mathbf{M}^{t+1} = \hat{\boldsymbol{\Sigma}}^{1/2} (\hat{\boldsymbol{\Sigma}}^{1/2} \boldsymbol{\Pi}^{t+1} \hat{\boldsymbol{\Sigma}}^{1/2} - \mathbf{H}^{t+1} + \boldsymbol{\Gamma}^{t+1}) \hat{\boldsymbol{\Sigma}}^{1/2}$;
    \State Broadcast $\boldsymbol{\Pi}^{t+1}$, $\mathbf{M}^{t+1}$ to clients; 
\EndWhile
\end{algorithmic}
\end{algorithm}

ADMM is an algorithm developed for optimization problems with the objective and constraint terms divided into different parts. 
This situation occurs in many statistical learning contexts, e.g., distributed machine learning with aggregated loss function, generalized $l_1$-norm regularization loss minimization. 
We observe that (\ref{eq2.10}) also has this separable property and can be solved by a linearized ADMM algorithm. 
Algorithm \ref{alg1} presents a detailed process of our method. 
In the $\boldsymbol{\Pi}$-update step, we first compute the solution $\boldsymbol{\Pi}_{i}^{t+}$ to the local problem distributedly on each client and then update $\boldsymbol{\Pi}^{t+1}$ by averaging the solution $\boldsymbol{\Pi}_{i}^{t+}$'s. 
The derivation of Algorithm \ref{alg1} is deferred to the appendix. 
Fang et al. \cite{Fang2015GADMM} studied the convergence of the linearized version of ADMM algorithms and established a worst-case $\cO(1/t)$ convergence rate, where $t$ is the iteration counter. 
We pick $K$ eigenvectors of $\hat{\boldsymbol{\Pi}}$ with the largest eigenvalues as the estimation $\hat{\mathbf{B}}$ of sufficient dimension reduction directions. 
The sparse model is estimated as $\{1 \leq j \leq d: \hat{\mathbf{B}}_{j \cdot} \neq \mathbf{0} \}$, where $\hat{\mathbf{B}}_{j \cdot}$ denotes the $j$-th row vector of $\hat{\mathbf{B}}$. 

\subsection{Hyper-parameter selection}
\label{sec2.3}
There are some hyper-parameters in our proposed method: $K$ the dimension of the central subspace and $\rho$ that controls the degree of sparsity. 
To determine the structural dimension of the central subspace, there are four main ways: sequential test, bootstrap, cross validation, and Bayesian information criterion (BIC), see \cite{Li2017SDR}. 
Constructing a sequential test, according to \cite{zhu2010cumulative}, is challenging due to the complicated structure of asymptotic variance and the diverging number of degrees of freedom. 
The next two methods both need multiple rounds of estimation and are too complicated for federated learning. 
Therefore, we consider using BIC to determine the dimension $\hat{K}_i$ on each client, and then take the mode of all $\hat{K}_i$'s as the final $\hat{K}$. 
If this produces two or more modes, we randomly choose one of them with equal probability. 
Following \cite{zhu2006sirhdc, zhu2010cumulative}, we define a BIC type criterion on client $i$ as follows 
\begin{equation}
    \label{eq2.14a} \hat{K}_i = \argmax_{k \in [d-1]} n_i \sum_{j = 1}^{k} \hat{\lambda}_{i, j}^2 / \sum_{j=1}^{d} \hat{\lambda}_{i, j}^2 - C_{n_i} k (k + 1) / 2, 
\end{equation} where $\hat{\lambda}_{i, j}$ is the $j$-th eigenvalue of $\hat{\mathbf{T}}_i$. 
Then our estimator $\hat{K}$ is the mode of the set $\{ \hat{K}_i \vert i \in [m] \}$. 
Zhu et al. \cite{zhu2010cumulative} proved that $\hat{K}_i$ converges to $K$ in probability, as $n_i \rightarrow \infty$, as long as $n_i^{-1} C_{n_i} \rightarrow 0$ and $C_{n_i} \rightarrow \infty$. 
We have the following corollary. 
\begin{corollary}
\label{cor:bic}
Assume the covariates on different clients are $d$-dimensional sub-Gaussian random vectors and $d = o((N/m)^{1/2})$, then $\hat{K}$ converges to $K$ in probability, if $n_i^{-1} C_{n_i} \rightarrow 0$ and $C_{n_i} \rightarrow \infty$, $i \in [m]$. 
\end{corollary}
The proof of Corollary \ref{cor:bic} and numerical results on dimension determination are referred to Appendix \ref{bic}. 
Throughout the context, we choose $C_n = n^{1/2} + 0.5 \log(n)$, which performs quite well in numerical studies. 

We use hold-out validation to select the tuning parameter $\rho$. 
Let $R(\bx)$ be the $K$-dimensional sufficient dimension reduction variable, compute the top $K$ eigenvectors $\hat{\boldsymbol{\pi}}_1, \dots, \hat{\boldsymbol{\pi}}_K$ of $\hat{\boldsymbol{\Pi}}$ and we can get an estimator $\hat{R}(\bx) = (\hat{\boldsymbol{\pi}}_1^{\mathrm{T}} \bx, \dots, \hat{\boldsymbol{\pi}}_K^{\mathrm{T}} \bx)^{\mathrm{T}}$. 
Following \cite{Kean2018convex}, the conditional expectation $\bE [y^{(i)} \vert \bx^{(i)} = \bx^{\prime}]$ can be estimated as
\begin{equation}
    \label{eq2.14}
    \hat{\bE} [y^{(i)} \vert \bx^{(i)} = \bx^{\prime}] = \frac{\sum_{j=1}^{n_i} y_j^{(i)} \exp \{-\frac{1}{2}\|\hat{R}(\bx^{\prime}) - \hat{R}(\bx_j^{(i)})\|_2^2\}}{\sum_{j=1}^{n_i} \exp\{-\frac{1}{2}\|\hat{R}(\bx^{\prime}) - \hat{R}(\bx_j^{(i)})\|_2^2\}}. 
\end{equation}
We first separate $S_i$ into training set $S_{i,tr}$ and validation set $S_{i,val}$ on each client and use Algorithm \ref{alg1} to get a solution $\hat{\boldsymbol{\Pi}}$ from $S_{i,tr}$'s. 
Then we can predict the estimated conditional mean for samples in $S_{i,val}$'s using (\ref{eq2.14}). 
Given a grid of $\rho$, we choose the best parameter to minimize the validation error $\sum_{i=1}^{m} \sum_{j \in S_{i,val}} (y_j^{(i)} - \hat{\bE}[y^{(i)} \vert \bx^{(i)} = \bx_j^{(i)}])^2/\lvert S_{i,val}\rvert$, where $\lvert S \rvert$ is the cardinality of the set $S$. 

\section{Theoretical Studies}
\label{theoretical}

In this section, we will discuss some theoretical properties of our method. 
For simplicity, suppose there are $m$ clients in the system and client $i$ has $n_i$ independent observations, which follow the distribution $\cD_i$, such that $(n_1, \dots, n_m) \sim \mathrm{Multinomial}(N, \omega_1, \dots, \omega_m)$. 
We also assume that the distributions $\cD_1, \dots, \cD_m$ are independent conditioned on $n_1, \dots, n_m$. 
The central subspace $\cV_{y \vert \bx}$ is assumed to have a sparsity level $$s = \lvert \mathrm{supp} \{\mathrm{diag}(\boldsymbol{\Pi})\} \rvert, $$where $\boldsymbol{\Pi} = \mathbf{V} \mathbf{V}^{\mathrm{T}}$ is the projection matrix onto $\cV_{y \vert \bx}$. 
$\mathrm{diag}(\cdot)$ denotes the operation that maps a matrix to a vector composed of its diagonal elements. 
$\mathrm{supp}(\mathbf{v})$ denotes the support set of a d-dimensional vector $\mathbf{v}$, which is defined as $\{ i \in [d] : \lvert \mathbf{v}_i \rvert > 0\}$. 
The sparsity level of the central subspace is the number of non-zero diagonal elements in the projection matrix. 
Suppose $\boldsymbol{\Pi}_{j j} = 0$, using the fact that $\boldsymbol{\Pi}_{j j} = \sum_{k=1}^{K} \mathbf{V}_{j k}^2$, and we have $\mathbf{V}_{j k} = 0$ for $k \in [K]$. 
This means when the $j$-th diagonal element of the projection matrix is zero, the entire $j$-th row of $\mathbf{V}$ is zero, indicating not to select the $j$-th variable. 

Our goal is to establish an upper error bound for the estimator $\hat{\boldsymbol{\Pi}}$ obtained from the FedSSIR algorithm under the non-asymptotic setting, where $N$, $m$, $d$, $K$, $s$ are allowed to grow. 
We first provide concentration inequalities on the estimators $\hat{\boldsymbol{\Sigma}}$ and $\hat{\mathbf{T}}$. 
Based on the concentration results, we can measure the distance between the population and estimated subspaces. 
Here are some regularity conditions. 
\begin{enumerate}[label=(C\arabic*)]
    \item The distribution $\cD_i$ satisfies the linear condition (\ref{eq2.4}), $\forall i \in [m]$. 
    \item The covariates on different clients are $d$-dimensional sub-Gaussian random vectors with a same covariance matrix $\boldsymbol{\Sigma}$. 
    \item The largest generalized eigenvalue $\lambda_1$ of matrices $\{ \bar{\mathbf{T}}, \boldsymbol{\Sigma} \}$ is bounded by some constant. 
    \item The inverse regression function $m^{i}(y^{(i)}) = \bE[\bx^{(i)} \vert y^{(i)}]$ has a bounded total variation, $\forall i \in [m]$. 
    \item The dimension of central subspace $K < \min(s, \log d)$. 
    \item There exists a constant $c > 0$ such that $1/c \leq \lambda_{\min} (\boldsymbol{\Sigma}, s) \leq \lambda_{\max} (\boldsymbol{\Sigma}, s) \leq c$, where $\lambda (\boldsymbol{\Sigma}, s)$ denotes the $s$-sparse eigenvalue of $\boldsymbol{\Sigma}$. 
\end{enumerate}

Condition (C1) is the commonly assumed linear condition in SIR literature. 
Condition (C2) states a sub-Gaussian tail decay of high dimensional random vectors, see \cite{vershynin2011introduction, vershynin2018HDP} for a comprehensive discussion of sub-Gaussian random variables. 
(C2) indicates that our theoretical results don't require the i.i.d. assumption of covariates as long as they are independent random sub-Gaussian vectors and have an identical covariance matrix. 
Condition (C3) is a mild condition on the generalized eigensystem. 
Condition (C4) is an assumption on the smoothness of the inverse regression curve. 
A similar assumption has been given in some other SIR literature, see \cite{hsing1992asym, zhu1995asymptotics, zhu2006sirhdc}. 
Condition (C5) states an assumption on the number of dimension reduction directions and the sparsity level $s$ of the central subspace. 
Condition (C6) is widely used in the high-dimensional literature, see \cite{Meinshausen2009lassohigh, zhang2010nearly} for example. 
The concepts of total variation and $s$-sparse eigenvalue $\lambda (\boldsymbol{\Sigma}, s)$ of $\boldsymbol{\Sigma}$ are given in the Appendix \ref{defs}. 

\begin{lemma}
\label{lem1}
Assume that Condition (C2) holds, then there exists constants $C_1, C_1^{'} > 0$ such that $$\|\hat{\boldsymbol{\Sigma}} - \boldsymbol{\Sigma} \|_{\max} \leq C_1 (\log d / N)^{1/2}, $$ with probability greater or equal to $1 - \exp (-C_1^{'} \log d)$. 
\end{lemma}
Lemma \ref{lem1} shows that the modified FedSVD estimator of $\boldsymbol{\Sigma}$ shares the same tail bound as the classic batch estimator \cite{ravikumar2011high}. 

To get the basis estimator $\hat{\mathbf{V}}$, we need to estimate the covariance matrix of the conditional expectation $\mathbf{T}_i$. 
Obviously, we have the identity $\mathbf{T}_i = \boldsymbol{\Sigma}_i - \mathbf{Q}_i$, where $\mathbf{Q}_i$ denotes the average covariance matrix $\bE[\cov(\mathbf{x}^{(i)} \vert y^{(i)})]$. 
Two consistent estimators for $\mathbf{Q}_i$ are given in Appendix \ref{defs}. 
\begin{lemma}
\label{lem2}
Assume that Conditions (C2)-(C4) and (C6) hold and the response $y^{(i)}$ is bounded for all clients. 
Also, $m \leq C_1 N^{\eta}$ for some $C_1$, where $\eta \in (0, 1)$. 
There exists constants $C_2, C_2^{\prime}, C_3, C_4 > 0 $ such that $$\|\hat{\mathbf{Q}} - \bar{\mathbf{Q}}\|_{\max} \leq C_2 (\log d / N)^{1/2} + C_3 (\log d)^{1/2} / N^{1 - \eta}, $$ with probability at least $1 - \exp(-C_4 \log d)$, where $\hat{\mathbf{Q}} = \sum_{i=1}^{m} \frac{n_i}{N} \hat{\mathbf{Q}}_i$ and $\bar{\mathbf{Q}} = \sum_{i=1}^{m} \omega_i \mathbf{Q}_i$. 

If we further assume that $m \leq C_1 N^{1/2}$, we have $$\|\hat{\mathbf{Q}} - \bar{\mathbf{Q}}\|_{\max} \leq C_2^{\prime} (\log d / N)^{1/2}, $$ with probability greater or equal to $1 - \exp(-C_4 \log d)$. 
\end{lemma}
It is shown in \cite{Kean2018convex} that when there is only one client, the estimation error of $\hat{\mathbf{Q}}$ is of order $(\log d / n)^{1/2}$ in the high-dimensional setting. 
Our result coincides with it under the assumption that the ratio of $m$ to $N$ is within an appropriate range. 
Directly from Lemma \ref{lem1} and Lemma \ref{lem2}, using the identities $\mathbf{T}_i = \boldsymbol{\Sigma}_i - \mathbf{Q}_i$ and $\hat{\mathbf{T}}_i = \hat{\boldsymbol{\Sigma}}_i - \hat{\mathbf{Q}}_i$, we have the following corollary. 
\begin{corollary}
\label{cor1}
Under conditions in Lemma \ref{lem1} and Lemma \ref{lem2}, assume that $m \leq C_1 N^{\eta}$ for some $C_1$, where $\eta \in (0, 1)$. 
There exists $C_2, C_2^{\prime}, C_3, C_4 > 0$ such that $$\|\hat{\mathbf{T}} - \bar{\mathbf{T}}\|_{\max} \leq C_2 (\log d / N)^{1/2} + C_3 (\log d)^{1/2} / N^{1 - \eta}, $$ with probability at least $1 - \exp(- C_4 \log d)$. 

Furthermore, if we assume that $m \leq C_1 N^{1/2}$, we have $$\|\hat{\mathbf{T}} - \bar{\mathbf{T}}\|_{\max} \leq C_2^{\prime} (\log d / N)^{1/2}, $$ with probability at least $1 - \exp(-C_4 \log d)$. 
\end{corollary}
Lemma \ref{lem1} and Corollary \ref{cor1} state that when the ratio of $m$ to $N$ is within an appropriate range, our estimations of covariance matrix and covariance matrix of the conditional expectation achieve the convergence rate proportional to $(\log d / N)^{1/2}$. 
This implies our estimation of the central subspace based on $\hat{\boldsymbol{\Sigma}}$ and $\hat{\mathbf{T}}$ can achieve the same error bound under certain conditions. 
Next, we state the theoretical result regarding the subspace distance between the central subspace and our estimation. 

The distance between the estimated subspace and the true subspace is defined as $D (\cV, \hat{\cV}) = \|\mathbf{P}_{\boldsymbol{\Pi}} - \mathbf{P}_{\hat{\boldsymbol{\Pi}}}\|_{\mathrm{F}}$, where $\mathbf{P}_{\boldsymbol{\Pi}}$ and $\mathbf{P}_{\hat{\boldsymbol{\Pi}}}$ are the projection matrices onto $\cV$ and $\hat{\cV}$. 
We can get the upper bound of the statistical error for our estimation. 
\begin{theorem}
\label{thm1}
Let $\cV$ and $\hat{\cV}$ be the true and estimated subspaces. 
Denote $\lambda_k$ as the k-th generalized eigenvalue of matrices $\{\hat{\mathbf{T}}, \hat{\boldsymbol{\Sigma}}\}$. 
Assume $N > C s^2 \log d/ \lambda_K^2$, $m \leq C' N^{\eta}$, $\eta \in (0, 1)$ for some constants $C$, $C'$ and the number of active covariates $s > \lambda_K K^2 / \log d$. 
Under Conditions (C1)-(C6), let $\rho = r (C_1 (\log d/ N)^{1/2} + C_1^{\prime} (\log d)^{1/2} / N^{1 - \eta})$ hold with some constants $C_1$, $C_1^{\prime}$, where $r \in [1, r_0]$, $r_0$ is a constant greater than 1, with probability at least $1 - \exp(-C_4 s) - \exp(-C_5 \log d)$, we have $$D (\cV, \hat{\cV}) \leq C_2 s (\log d/ N)^{1/2} / \lambda_K + C_3 s (\log d)^{1/2} / (\lambda_K N^{1 - \eta}), $$especially when $m \leq C^{\prime} N^{1/2}$, we have $$D (\cV, \hat{\cV}) \leq C_2^{\prime} s (\log d/ N)^{1/2} / \lambda_K, $$ for some constants $C_2$, $C_2^{\prime}$, $C_3$, $C_4$ and $C_5$. 
\end{theorem}

As we can see, under appropriate conditions on $m$, $N$ and $d$, estimated subspace $\hat{\cV}$ can achieve the statistical error rate of order $s (\log d/ N)^{1/2} / \lambda_K$ with high probability. 
When $m = 1$, the statistical error rate reduces to $s (\log d/ n)^{1/2} / \lambda_K$, which coincides with traditional sparse SIR. 
This shows that our approach can make use of data from all clients to get better results than a single client, solving the problem of insufficient data on a single client. 
Our results say that when the sample size of a single client is not too small, compared with the number of clients in the federated learning system, FedSSIR under non-i.i.d. setting can achieve the same statistical error rate as whole sample SIR under i.i.d. setting.

We conclude this section with a remark on the federated learning setting. 
\begin{remark}
Note that we allow $m$, the number of clients in the federated learning system, to approach infinity at a rate close to $N$, the total sample size of all clients. 
The average sample size $n = N / m$ on each client is roughly of order $N^{1 - \eta}$. 
And in our federated setting, $m$ doesn't have a direct relationship with $n$. 
When $\eta > 1/2$, it comes to a situation where $m$ tends to infinity faster than $n$, which also occurs in federated settings. 
For example, McMahan et al. \cite{McMahan2016} proposed federated learning as a solution for updating language models on mobile phones. 
There are many mobile edge devices holding private data while the training sample size of each device may not be enough. 
Theorem \ref{thm1} says our method remains effective in this case. 
\end{remark}

\section{Numerical Studies}
\label{numerical}

\subsection{Simulation Studies}

We evaluate the performance of our method through simulations. 
The simulation datasets are generated using the following models. 
\begin{align*}
    \text{Model 1: } & y = \boldsymbol{\beta}_1^{\mathrm{T}} \bx + \varepsilon, \\
    \text{Model 2: } & y = \exp(\boldsymbol{\beta}_1^{\mathrm{T}} \bx / 3^{1/2} + \varepsilon), \\
    \text{Model 3: } & y = \frac{\boldsymbol{\beta}_1^{\mathrm{T}} \bx}{0.5 + (\boldsymbol{\beta}_2^{\mathrm{T}} \bx + 1.5)^2} + \varepsilon, \\
    \text{Model 4: } & y = \sgn(\boldsymbol{\beta}_1^{\mathrm{T}} \bx) \times \lvert 2 + \boldsymbol{\beta}_2^{\mathrm{T}} \bx \rvert^{-1} + \varepsilon. \\
\end{align*}
We set $\boldsymbol{\beta}_{1,j} = 1$ for $j = 1, 2, 3$ and $\boldsymbol{\beta}_{1,j} = 0$ otherwise; 
$\boldsymbol{\beta}_{2,j} = 1$ for $j = 4, 5$ and $\boldsymbol{\beta}_{2,j} = 0$ otherwise. 

We use two different ways to generate datasets for federated learning. 
The first one is a modified version of the federated learning benchmark LEAF \citep{Caldas2019leaf}. 
For client $i \in [m]$: 
\begin{enumerate}
    \item Generate $\mathbf{A}_i \sim N(0, \alpha \mathbf{I}_d)$; 
    \item Generate $\mathbf{v}_i \sim N(\mathbf{A}_i, \mathbf{I}_d)$; 
    \item Draw $\bx^{(i)}$ from $N(\mathbf{v}_i, \boldsymbol{\Sigma})$, where $\boldsymbol{\Sigma}$ is the covariance matrix with $\boldsymbol{\Sigma}_{i,j} = \gamma^{\lvert i-j \rvert}$; 
    \item Generate $\varepsilon \sim N(0, 1)$ and compute $y^{(i)}$ with respect to the models. 
\end{enumerate}
In our simulations, we choose $\alpha$ to be $1$ and $\gamma = 0.5$. 
This approach reflects the \textit{covariate shift} in section \ref{sec2.1}, which corresponds to different marginal distributions of $\bx^{(i)}$'s. 
Simulation datasets of Setting 1-4 are generated in this way with respect to Model 1-4. 

Another one pays attention to the \textit{concept shift}, which focuses on the gap of the conditional distribution of $y$ given $\bx$. 
For the $i$-th client, $\bx^{(i)}$ is generated from a normal distribution with zero mean and the same covariance matrix structure as above. 
However, we assume that $y^{(i)}$ on different clients may follow different models, which means we don't know whether the dataset on a client comes from Model 1 or Model 2 in Setting 5 and Model 3 or Model 4 in Setting 6. 
Thus, we generate a Bernoulli random variable $b_i \sim B(0.5)$ for each client to decide which model $y^{(i)}$ belongs to. 
In Setting 5, if $b_i = 1$, then the $i$-th client follows Model 1, otherwise follows Model 2; 
in Setting 6, if $b_i = 1$, then the $i$-th client follows Model 3, otherwise follows Model 4. 

Therefore, we have six settings of simulation experiments. 
The goal is to estimate the central subspace $\cV_{y \vert \bx}$ using $\text{span}(\hat{\boldsymbol{\Pi}})$. 
Also, we want to measure the variable selection accuracy of our algorithm. 
We use the definition proposed by Tan et al. \cite{Kean2018convex} that the true positive rate (TPR) is the proportion of correctly identified variables, and the false positive rate (FPR) is the proportion of inactive variables falsely identified as active variables. 

\begin{table}[t]
\begin{center}
\begin{minipage}{\textwidth}
\caption{True and false positive rates, and subspace distances with $m = 10$, $n = 100$ and $200$, $d = 150$. 
All entries are averaged across 200 runs. 
The standard deviations are in the brackets. 
}
\label{tab1}
\begin{tabular}{@{}cccccccc@{}}
\toprule
                                &      & Setting 1 & Setting 2 & Setting 3 & Setting 4 & Setting 5 & Setting 6 \\ \midrule
$n = 100$ &      & & & & & & \\
FedSSIR		& TPR               & 1.000     & 1.000	 & 0.946 & 0.897 & 1.000	  & 0.958 \\ 
            &                   & (0)       & (0)   & (0.128)&(0.180)& (0)      & (0.101) \\ 
								& FPR & 0.002 & 0.006 & 0.012 & 0.006 & 0.003 & 0.014 \\ 
            &                   & (0.003) & (0.007) & (0.009) & (0.007) & (0.004) & (0.011) \\ 
								& Dist & 0.088 & 0.167 & 0.630 & 0.862 & 0.105 & 0.654 \\ 
            &                   & (0.044) & (0.201) & (0.324) & (0.371) & (0.048) & (0.302) \\ 
SSIR         	& TPR           & 0.917 & 0.803 & 0.453 & 0.160 & 1.000     & 1.000     \\ 
            &                   & (0.236) & (0.341) & (0.336) & (0.276) & (0) & (0) \\ 
								& FPR  & 0.347 & 0.280 & 0.199 & 0.046 & 0.129 & 0.075 \\ 
            &                   & (0.321) & (0.304) & (0.228) & (0.105) & (0.145) & (0.201) \\ 
								& Dist & 1.755 & 1.514 & 1.864 & 1.709 & 0.892 & 0.790 \\ 
            &                   & (0.175) & (0.312) & (0.178) & (0.162) & (0.475) & (0.303) \\ 
LassoSIR      & TPR             & 1.000     & 1.000 	 & 0.789 & 0.686 & 1.000     & 0.972 \\ 
            &                   & (0) & (0) & (0.162) & (0.171) & (0) & (0.078) \\ 
								& FPR  & 0.143 & 0.110 & 0.172 & 0.128 & 0.058 & 0.127 \\ 
            &                   & (0.067) & (0.056) & (0.068) & (0.068) & (0.053) & (0.091) \\ 
								& Dist & 0.176 & 0.190 & 1.198 & 1.353 & 0.142 & 0.998 \\ 
            &                   & (0.076) & (0.082) & (0.110) & (0.171) & (0.062) & (0.156) \\
\midrule
$n = 200$ &      & & & & & & \\
FedSSIR			& TPR  & 1.000	  & 1.000	 & 0.962 & 0.975 & 1.000	  & 0.958 \\
                                            &&(0) & (0) & (0.116) & (0.092) & (0) & (0.111) \\ 
									& FPR  & 0.000 & 0.002 & 0.011 & 0.009 & 0.001 & 0.017 \\
									        &&(0.001) & (0.003) & (0.008) & (0.007) & (0.002) & (0.011) \\ 
									& Dist & 0.052 & 0.084 & 0.587 & 0.578 & 0.058 & 0.530 \\ 
									        &&(0.025) & (0.076) & (0.238) & (0.273) & (0.032) & (0.251) \\ 
SSIR         		& TPR  & 0.998 & 0.995 & 0.504 & 0.232 & 1.000 & 1.000 \\
                                            &&(0.024) & (0.053) & (0.360) & (0.341) & (0.000) & (0.000) \\ 
									& FPR  & 0.874 & 0.834 & 0.248 & 0.061 & 0.198 & 0.072 \\
									        &&(0.092) & (0.122) & (0.274) & (0.125) & (0.225) & (0.188) \\ 
									& Dist & 1.895 & 1.672 & 1.967 & 1.684 & 0.915 & 0.742 \\ 
									        &&(0.099) & (0.231) & (0.151) & (0.168) & (0.462) & (0.274) \\ 
LassoSIR         	& TPR  & 1.000     & 1.000     & 0.905 & 0.805 & 1.000     & 1.000 \\
                                            &&(0) & (0) & (0.142) & (0.165) & (0) & (0.000) \\ 
									& FPR  & 0.231 & 0.168 & 0.249 & 0.179 & 0.053 & 0.187 \\
									        &&(0.116) & (0.089) & (0.088) & (0.075) & (0.055) & (0.087) \\ 
									& Dist & 0.172 & 0.172 & 1.150 & 1.263 & 0.086 & 0.549 \\ 
									        &&(0.119) & (0.101) & (0.087) & (0.166) & (0.039) & (0.115) \\ 
\botrule
\end{tabular}
\end{minipage}
\end{center}
\end{table}

On each client, we need to estimate the covariance matrix of the conditional expectation. 
We use the sample covariance matrix and (\ref{eq2.6a}) to give the estimation, where each slice set the sample size of $n_h = 20$. 
The number of dimension reduction directions $K$ is chosen by the BIC-type criterion. 
The penalization parameter $\rho$ is selected by the hold-out validation outlined in section \ref{sec2.3}. 
For comparison, we include the results of sparse SIR (SSIR) \cite{Kean2018convex} and LassoSIR \cite{Lin2019LassoSIR} on the aggregated dataset. 
The tuning parameters of these two methods are selected by cross validation. 
We randomly repeat these experiments $200$ times with equal sample size $n = 100$ or $200$ on $m = 10$ clients and covariate dimension $d = 150$. 

The results are summarized in Table \ref{tab1}. 
Our proposal yields a better estimation accuracy in the non-i.i.d. federated data. 
Our method also outperforms other methods in terms of true and false positive rates. 
It seems that SSIR and LassoSIR may not be suitable for the \textit{covariate shift}. 
FedSSIR has good performance in the high-dimensional case that $n = 100$. 
Comparing the results of $n = 100$ with $n = 200$, we can see that with the increase of $n$, our method improves in TPR, FPR and estimation error, which is also consistent with our theoretical results. 
The results show that our method is robust under non-i.i.d. settings. 

\begin{figure}[htbp]
	\begin{minipage}[t]{0.3\linewidth}
		\centering
		\includegraphics[scale=0.18]{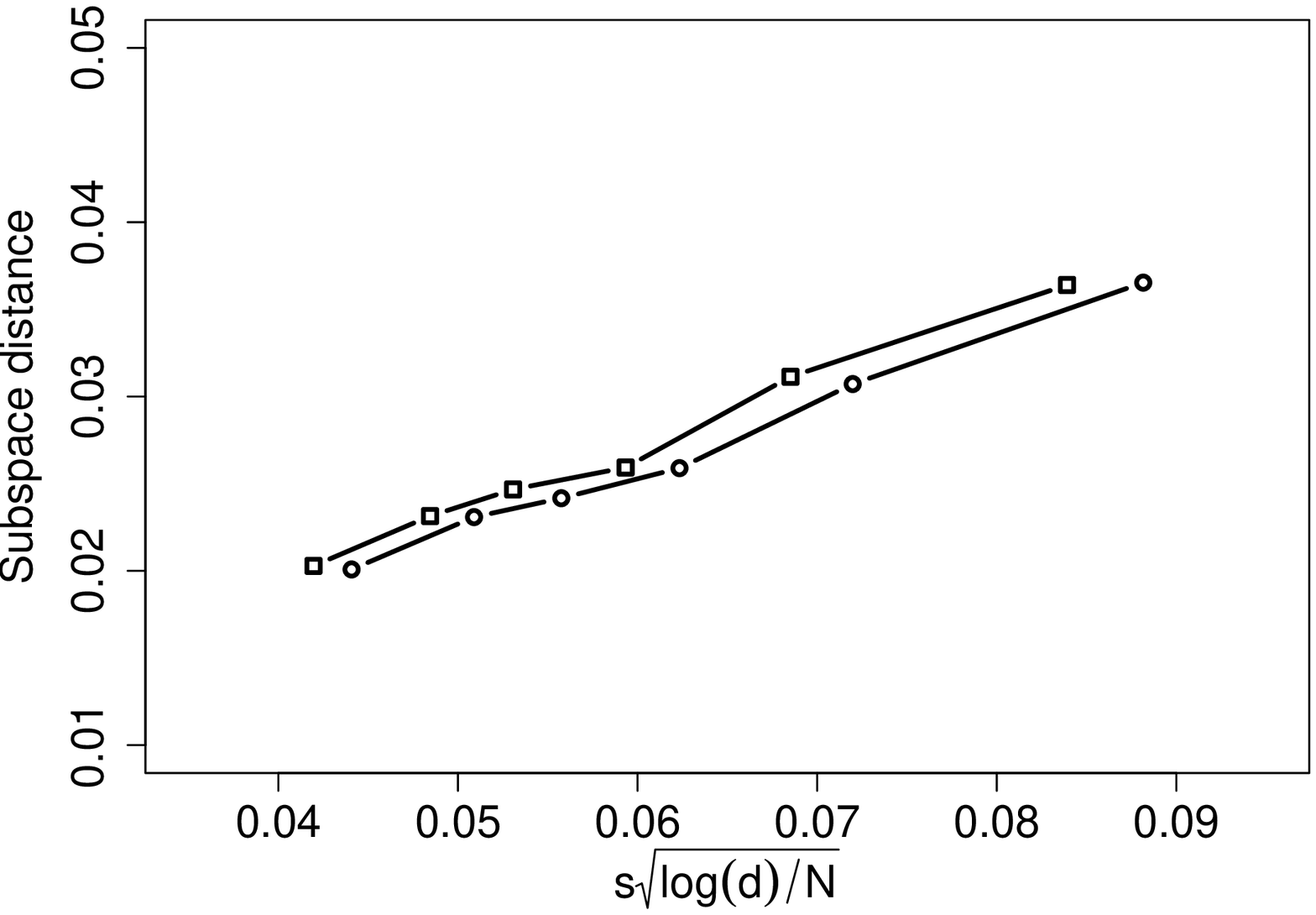}
		\caption{Results for the sub-space distance, averaged on 500 data sets of Setting 1. \label{fig:1}}
	\end{minipage}
	\begin{minipage}[t]{0.3\linewidth}
		\centering
		\includegraphics[scale=0.18]{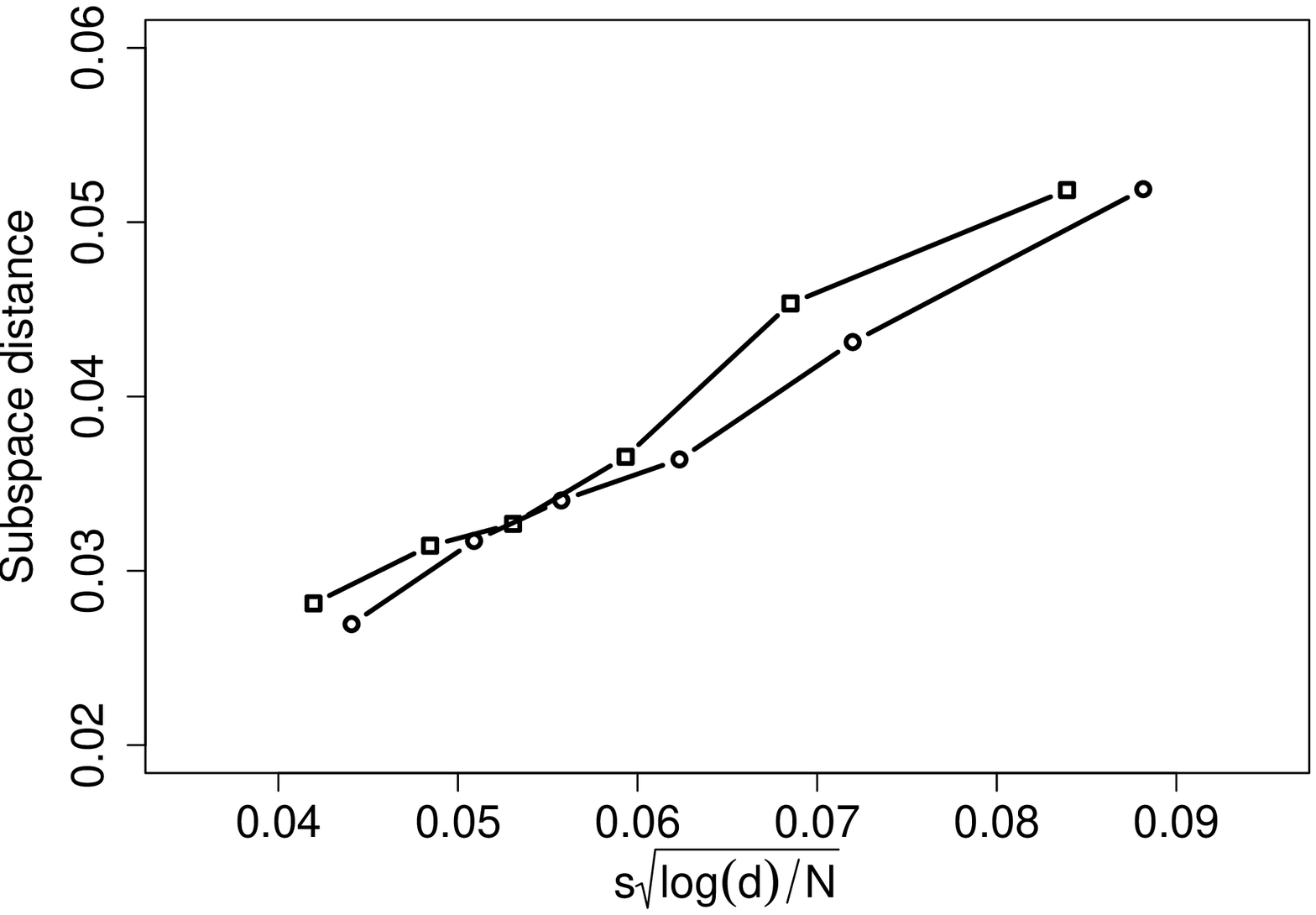}
		\caption{Results for the sub-space distance, averaged on 500 data sets of Setting 2. \label{fig:2}}
	\end{minipage}
	\begin{minipage}[t]{0.3\linewidth}
		\centering
		\includegraphics[scale=0.18]{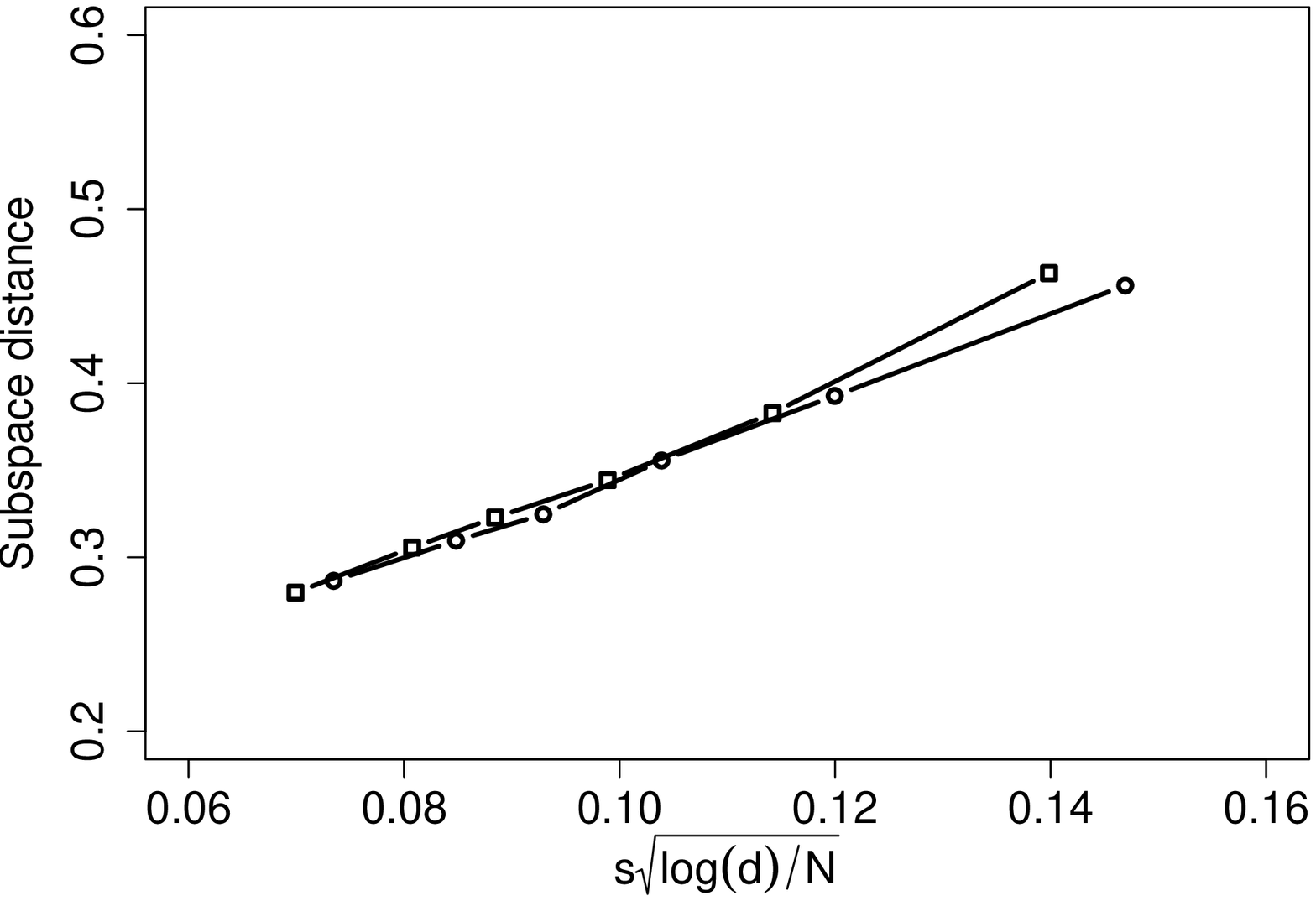}
		\caption{Results for the sub-space distance, averaged on 500 data sets of Setting 3. \label{fig:3}}
	\end{minipage}
\end{figure}

\begin{figure}[htbp]
	\begin{minipage}[t]{0.3\linewidth}
		\centering
		\includegraphics[scale=0.18]{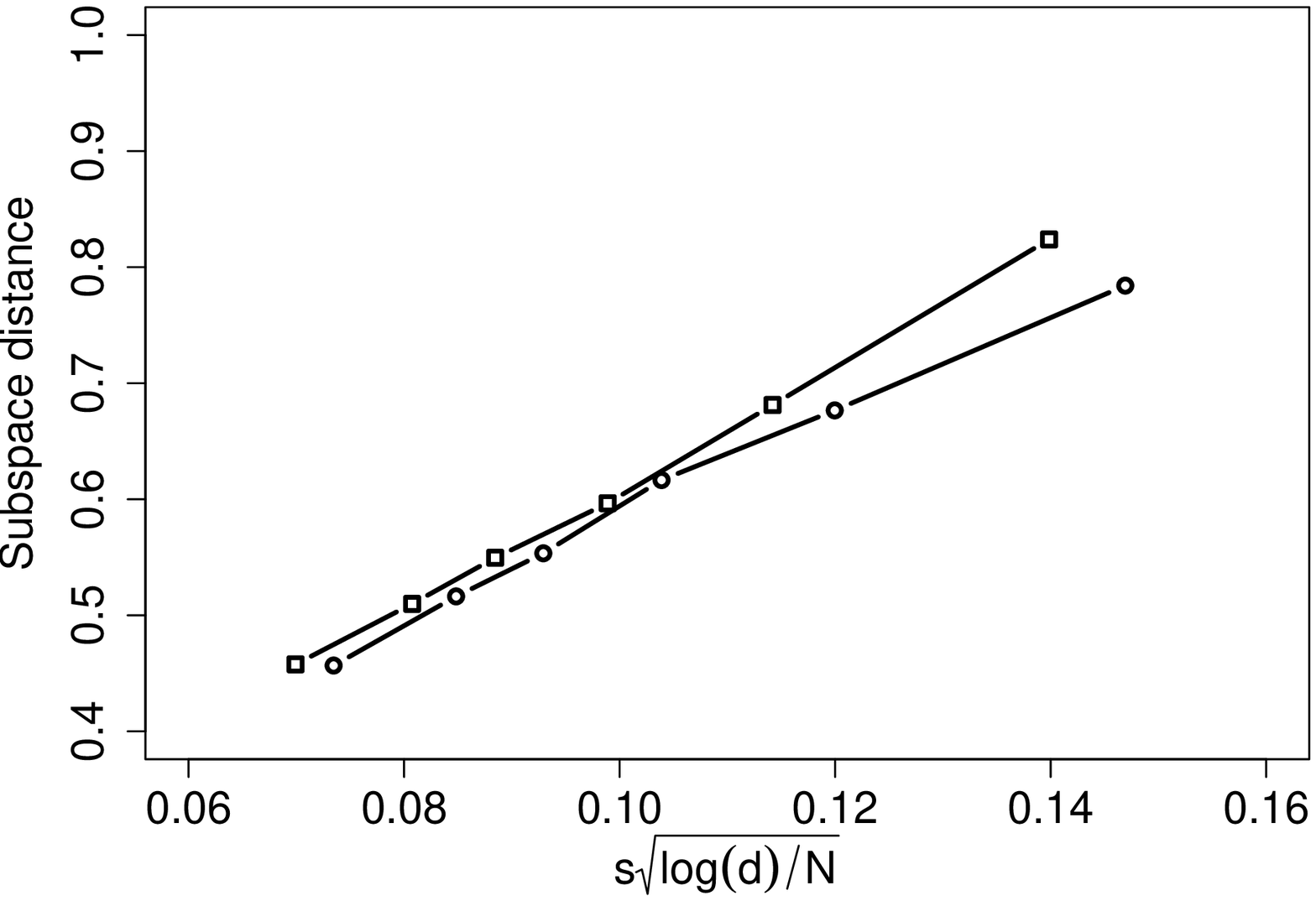}
		\caption{Results for the sub-space distance, averaged on 500 data sets of Setting 4. \label{fig:4}}
	\end{minipage}
	\begin{minipage}[t]{0.3\linewidth}
		\centering
		\includegraphics[scale=0.18]{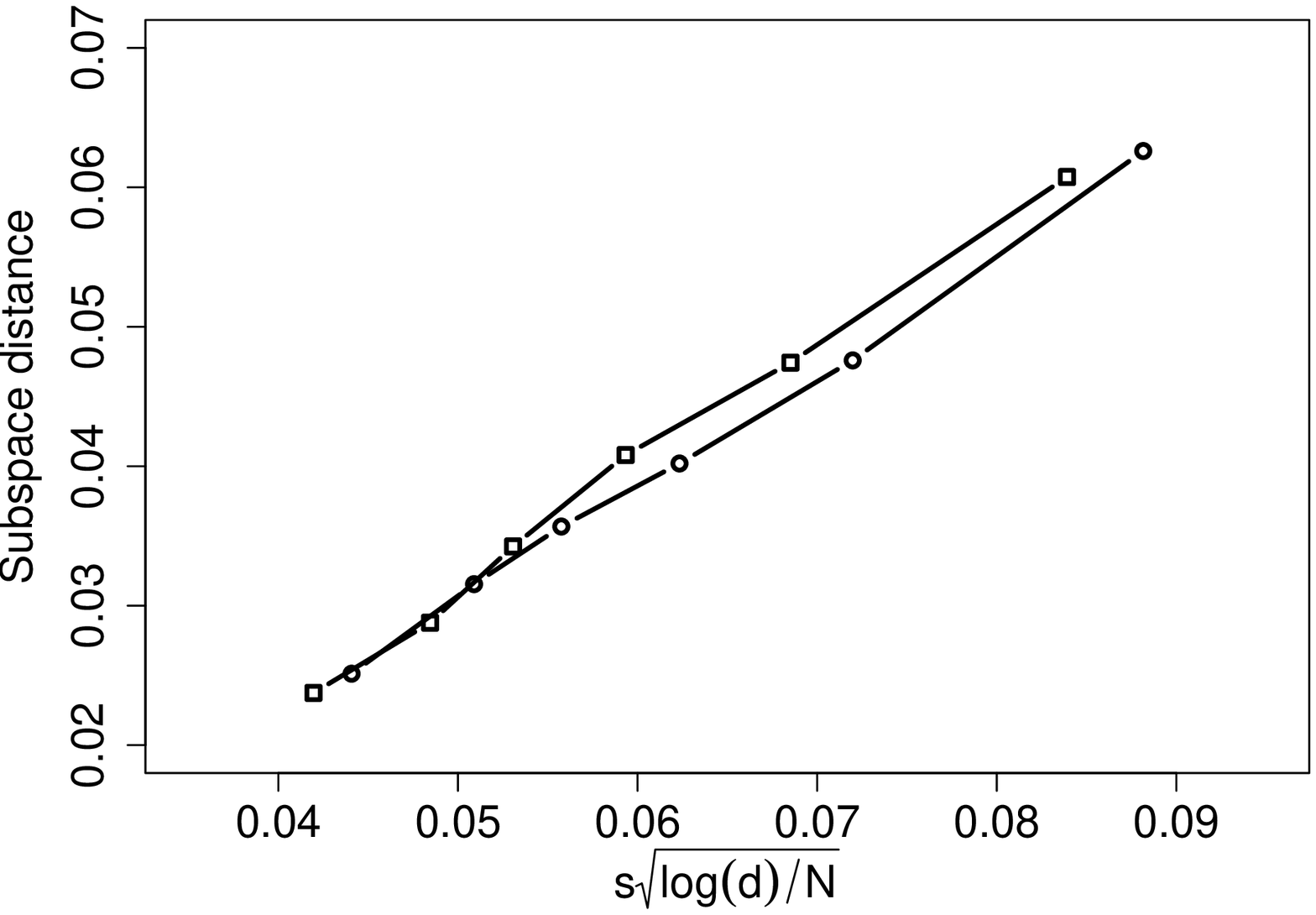}
		\caption{Results for the sub-space distance, averaged on 500 data sets of Setting 5. \label{fig:5}}
	\end{minipage}
	\begin{minipage}[t]{0.3\linewidth}
		\centering
		\includegraphics[scale=0.18]{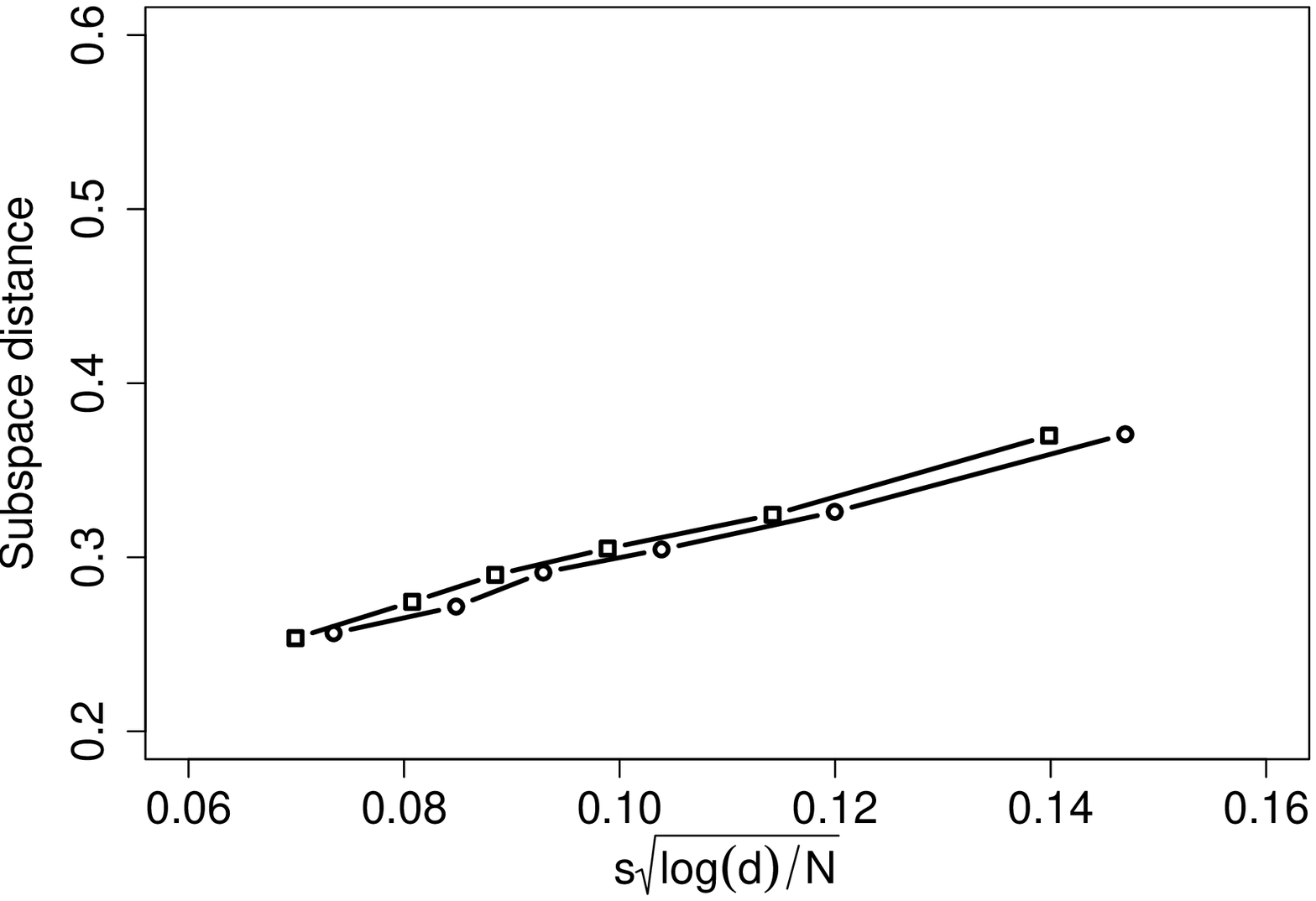}
		\caption{Results for the sub-space distance, averaged on 500 data sets of Setting 6. \label{fig:6}}
	\end{minipage}
\end{figure}

Next, we conduct experiments on simulated datasets to evaluate the statistical error rate. 
We employed the same data generation technique above except for one difference, where client sample sizes are unbalanced. 
We generate a random vector $\boldsymbol{\omega} \sim \mathrm{Dirichlet}_m(5)$ and $(n_1, \dots, n_m) \sim \mathrm{Multinomial}(N, \boldsymbol{\omega})$ with $m = 10$. 
In this case, we assume that the real $K$ is known and set $\rho = 0.5 (\log d / N)^{1/2}$. 
The estimation errors at different $N$ and $d$ are presented in Figure \ref{fig:1}-\ref{fig:6}. 
Circles represent results of $d = 75$ and squares represent results of $d = 50$. 
The figures demonstrate the subspace distance increases with respect to $s(\log d / N)^{1/2}$. 
This is consistent with Theorem \ref{thm1} where the statistical error has an order of $\cO(\sqrt{\frac{\log d}{N}})$. 


\begin{figure}[htbp]
    \centering
    \includegraphics[scale = 0.6]{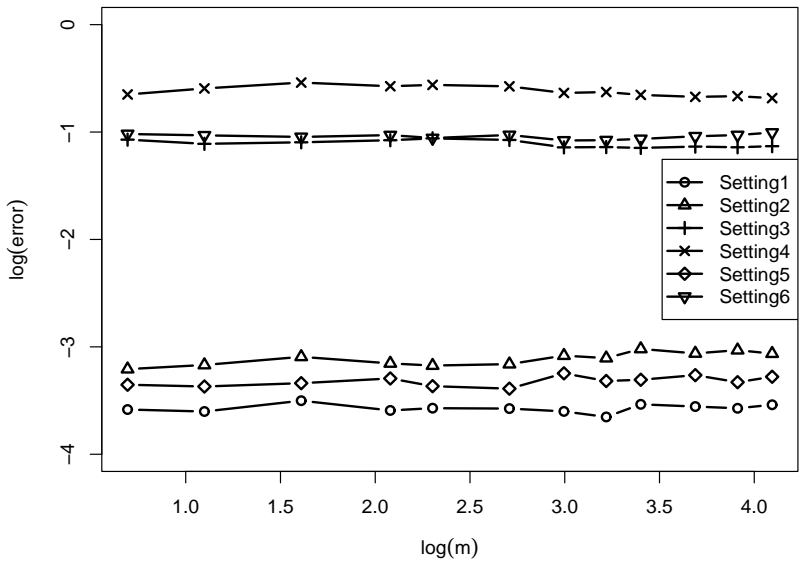}
    \caption{Statistical error with respect to the number of clients when the total sample
size $N = 10000$ is fixed. }
    \label{fig:7}
\end{figure}

We further investigate the performance of our method with the number of clients $m$ increasing. 
We keep $N = 10000$, $d = 100$ and split the data onto $m$ clients using the same heterogeneous setup. 
Figure \ref{fig:7} plots the subspace distance $D(\cV, \hat{\cV})$ against $m$. 
Each point on the plot is averaged on 200 replications. 
When the subsample size is small, or the number of clients is large, there is a slightly growing error of FedSSIR. 
This aligns with Condition (C4) and Lemma \ref{lem2} that it requires $n_i$ to be not too small to achieve optimal statistical performance. 
We can see that FedSSIR can efficiently estimate the central subspace as long as $m$ is in a reasonable range. 
Our method has stable performance under different settings. 

In order to analysis the impact of different $\omega_i$ on the simulation results, we conduct a sensitivity experiment against the mixture weight. 
We keep $N = 2000$, $d = 100$ and split the data onto $m$ clients using the heterogeneous set up. 
We generate a random vector $\boldsymbol{\omega} \sim \mathrm{Dirichlet}_m(\alpha)$ and $(n_1, \dots, n_m) \sim \mathrm{Multinomial}(N, \boldsymbol{\omega})$ with $m = 10$. 
The smaller the $\alpha$, the more extreme the $\boldsymbol{\omega}$. 
The number of dimension reduction directions $K$ is chosen by the BIC-type criterion. 
The penalization parameter $\rho$ is selected by the hold-out validation outlined in section \ref{sec2.3}. 
The results in Table \ref{tabomega} show that our method performs robustly against the choices of the mixture weight $\omega_i$. 
Our proposal has good performance for datasets with unbalanced sample size. 

\begin{table}[ht]
\begin{center}
\begin{minipage}{\textwidth}
\caption{True and false positive rates, and subspace distances with $\alpha = 1, 2, 5$, $N = 2000$, $m = 10$, $d = 100$. 
All entries are averaged across 200 runs. 
The standard deviations are in the brackets. 
}
\label{tabomega}
\begin{tabular}{@{}cccccccc@{}}
\toprule
            &      & Setting 1 & Setting 2 & Setting 3 & Setting 4 & Setting 5 & Setting 6 \\ \midrule
$\alpha=1$  & TPR  & 1.000 & 1.000 & 0.990 & 1.000 & 1.000 & 0.994 \\ 
            &      & (0) & (0) & (0.056) & (0) & (0) & (0.044) \\ 
			& FPR  & 0.000 & 0.001 & 0.011 & 0.014 & 0.000 & 0.012 \\ 
            &      & (0.001) & (0.003) & (0.004) & (0.006) & (0.002) & (0.005) \\ 
			& Dist & 0.033 & 0.044 & 0.419 & 0.388 & 0.033 & 0.417 \\ 
            &      & (0.016) & (0.022) & (0.132) & (0.114) & (0.018) & (0.097) \\
$\alpha=2$  & TPR  & 1.000 & 1.000 & 0.996 & 1.000 & 1.000 & 1.000 \\ 
            &      & (0) & (0) & (0.040) & (0) & (0) & (0) \\ 
			& FPR  & 0.000 & 0.001 & 0.011 & 0.013 & 0.000 & 0.012 \\ 
            &      & (0.001) & (0.002) & (0.004) & (0.005) & (0.001) & (0.003) \\ 
			& Dist & 0.031 & 0.041 & 0.388 & 0.399 & 0.034 & 0.386 \\ 
            &      & (0.016) & (0.020) & (0.093) & (0.086) & (0.018) & (0.062) \\ 
$\alpha=5$  & TPR  & 1.000 & 1.000 & 1.000 & 1.000 & 1.000 & 1.000 \\ 
            &      & (0) & (0) & (0) & (0) & (0) & (0) \\ 
			& FPR  & 0.000 & 0.000 & 0.011 & 0.013 & 0.000 & 0.011 \\ 
            &      & (0) & (0.002) & (0.002) & (0.005) & (0) & (0.003) \\ 
			& Dist & 0.028 & 0.040 & 0.376 & 0.385 & 0.032 & 0.370 \\ 
            &      & (0.015) & (0.020) & (0.072) & (0.075) & (0.017) & (0.042) \\
\botrule
\end{tabular}
\end{minipage}
\end{center}
\end{table}

In addition to the simulations above, we have carried out simulations measuring the performance of our method on non-Gaussian random variables. 
The results of this experiment are presented in Appendix \ref{bic}, along with the results of the previous experiments on BIC dimension determination. 

\subsection{Real Data Analysis}

Default of credit card clients dataset~\citep{YEH2009dccc} contains 30,000 samples of people with different credit risks. 
For each subject, there are 23 attributes. 
First, we separate this dataset into different clients using a heterogeneous pattern, where sample sizes and class proportions on different clients are unbalanced. 
For class $j$, we generate a random vector $\boldsymbol{\omega}_j$ with respect to a Dirichlet distribution $\mathrm{Dirichlet}_{m}(\alpha)$, where $\alpha$ is the concentration parameter that controls the heterogeneity of federated data. 
Smaller $\alpha$ yields more heterogeneous data. 
Then we generate a random vector $\mathbf{n}_{j}$ from the multinomial distribution $\mathrm{Multinomial}(\mathbf{\boldsymbol{\omega}}_{j})$. 
We allocate $\mathbf{n}_{j, i}$ samples of class $j$ to client $i$. 
Then we partition the samples on each client into a training set and a test set randomly of the same size. 

We use logistic regression to construct the classifier and take the average prediction accuracy in the test sets as the testing metric. 
We compare the performance of the following competitors. 
\begin{enumerate}[label=(M\arabic*)]
    \item Logistic regression on the aggregated training set. This serves as a benchmark for comparison. 
    \item Logistic regression on each client. 
    \item Federated sparse sliced inverse regression followed by logistic regression on dimension reduction variables on each client. 
    \item Federated principal component analysis by Grammenos et al. \cite{grammenos2020fpca} followed by logistic regression on dimension reduction variables on each client. 
    \item Federated principal component analysis by Chai et al. \cite{chai2021fedsvd} followed by logistic regression on dimension reduction variables on each client. 
\end{enumerate}
We choose the number of the dimension reduction directions $K = 1$ for all dimension reduction methods, and the penalization parameter $\rho$ of our method is chosen by cross validation. 
We repeat this process 100 times to obtain the mean performances of each method and standard deviations. 
The results are shown in Table~\ref{tab3}. 
It is clearly seen that FedSSIR (M3) performs better than its competitors (M4) and (M5). 
Also, FedSSIR performs better than local learner (M2) as the number of clients rises, which shows the significance of collaboration. 
In addition, FedSSIR is comparable to the global model (M1). 
This indicates that our method successfully captures the low dimensional structure. 

\begin{table}[ht]
\begin{minipage}{\textwidth}
\caption{The averaged testing accuracies based on 100 repetitions. The standard deviations are in the brackets. All entries are multiplied by 100. }
\label{tab3}
\begin{center}
\begin{tabular}{@{}ccccccc@{}}
\toprule
m & $\alpha$ & (M1) & (M2) & (M3) & (M4) & (M5) \\ \midrule
5 & 1 & 79.42(0.77) & 79.18(1.78) & 79.47(1.64) & 74.77(2.91) & 74.77(2.91) \\ 
5 & 2 & 79.71(0.70) & 79.86(1.45) & 79.96(1.47) & 75.22(2.51) & 75.22(2.51) \\ 
5 & 5 & 80.23(0.52) & 80.68(0.86) & 80.82(0.81) & 76.50(1.48) & 76.50(1.48) \\ 
10 & 1 & 79.37(0.60) & 78.66(1.46) & 79.17(1.45) & 74.51(2.34) & 74.51(2.34) \\ 
10 & 2 & 79.66(0.57) & 79.13(1.11) & 79.63(1.15) & 74.55(1.74) & 74.55(1.74) \\ 
10 & 5 & 80.20(0.37) & 80.24(0.63) & 80.63(0.63) & 76.32(1.06) & 76.33(1.06) \\ 
20 & 1 & 79.19(0.45) & 77.55(1.20) & 78.83(1.10) & 73.86(1.75) & 73.87(1.76) \\ 
20 & 2 & 79.56(0.42) & 78.46(1.01) & 79.54(0.83) & 74.71(1.54) & 74.71(1.54) \\ 
20 & 5 & 80.23(0.38) & 79.64(0.61) & 80.52(0.57) & 76.11(0.92) & 76.11(0.92) \\ 
30 & 1 & 79.13(0.37) & 77.06(1.11) & 78.99(0.98) & 73.93(1.60) & 73.94(1.59) \\ 
30 & 2 & 79.56(0.35) & 77.74(0.84) & 79.50(0.77) & 74.65(1.18) & 74.65(1.18) \\ 
30 & 5 & 80.19(0.33) & 78.92(0.55) & 80.50(0.42) & 75.98(0.68) & 75.98(0.68) \\
\botrule
\end{tabular}
\end{center}
\end{minipage}
\end{table}

The other one is the Communities and Crime dataset~\citep{misc_communities_and_crime_183}. 
It contains 1,944 observations of communities and 100 predictive features. 
The task is to predict the total number of violent crimes per 100K population of each sample. 
We first separate the communities into different clients by their states and select clients with sample size larger than 80 for our experiment. 
Thus we have 7 clients and a total of 1066 samples. 
Then we partition the samples on each client into a training set and a test set randomly of the same size. 

We use linear regression to construct the model and take the relative prediction error to evaluate the prediction performance, i.e. $\sum_{i \in test set} (\hat{y}_i - y_i)^2 / (y_i - \bar{y})^2$. 
We compare the performance of the following competitors. 
\begin{enumerate}[label=(M\arabic*)]
    \item Linear regression on the aggregated training set. This serves as a benchmark for comparison. 
    \item Linear regression on each client. 
    \item Federated sparse sliced inverse regression followed by linear regression on dimension reduction variables on each client. 
    \item Federated principal component analysis by Grammenos et al. \cite{grammenos2020fpca} followed by linear regression on dimension reduction variables on each client. 
    \item Federated principal component analysis by Chai et al. \cite{chai2021fedsvd} followed by linear regression on dimension reduction variables on each client. 
\end{enumerate}
We choose the number of the dimension reduction directions by BIC-type criterion, and the penalization parameter $\rho$ of our method is chosen by validation. 
We repeat this process 100 times to obtain mean performances of each method and standard deviations. 
The results are shown in Table~\ref{tab4}. 
FedSSIR (M3) is significantly superior to the unsupervised learners (M4)-(M5) in regression problems. 
Local learner (M2) performs badly in this case, which indicates that training with only a single client suffers from lack of data and results in under-fitting. 
Our FedSSIR method helps in this case, like other federated learning methods. 
Moreover, FedSSIR performs better than global learner (M1), which shows that our proposal captures the low dimensional structure and enhances the prediction power in the high dimensional case. 

\begin{table}[ht]
\begin{minipage}{\textwidth}
\caption{The averaged prediction errors based on 100 repetitions. The standard deviations are in the brackets. }
\label{tab4}
\begin{center}
\begin{tabular}{@{}ccccc@{}}
\toprule
 (M1) & (M2) & (M3) & (M4) & (M5) \\ \midrule
 0.355(0.024) & 2.059(0.033) & 0.312(0.023) & 0.400(0.022) & 0.398(0.022) \\ 
\botrule
\end{tabular}
\end{center}
\end{minipage}
\end{table}

\section{Discussions \& Conclusions}
\label{conclusion}
Sufficient dimension reduction in the context of federated learning is an exciting topic that hasn't gained much attention. 
We devised the FedSSIR algorithm to estimate the sufficient dimension reduction directions in the federated setting. 
With a $L_1$-norm penalty term, our method can simultaneously perform dimension reduction and variable selection in the high-dimensional setting. 
Our simulations and real data experiments show that FedSSIR can perform quite well on heterogeneous data. 
Also, this approach can easily be extended to other federated learning problems involving generalized eigenvalue decomposition. 
Nowadays, sufficient dimension reduction on a stochastic data stream is an essential topic in the big data era, see \cite{cai2020online, cheng2021online}, and how to establish a valid federated sufficient dimension reduction approach on several data streams is also an interesting problem. 

\clearpage

\begin{appendices}

\section{Derivation of Algorithm \ref{alg1}}

In this section, we derive the ADMM algorithm for solving (\ref{eq2.10}). 

We can write (\ref{eq2.10}) in the ADMM form
\begin{equation}
    \label{eq2.11}
    \min_{\boldsymbol{\Pi}, \mathbf{H} \in \cM} f(\boldsymbol{\Pi}) + g(\mathbf{H}) \text{ subject to } \hat{\boldsymbol{\Sigma}}^{1/2} \boldsymbol{\Pi} \hat{\boldsymbol{\Sigma}}^{1/2} = \mathbf{H}, 
\end{equation}
where $f(\boldsymbol{\Pi}) = \sum_{i=1}^{m} \frac{n_i}{N} (- \tr \{ \hat{\mathbf{T}}_i \boldsymbol{\Pi} \} + \rho \| \boldsymbol{\Pi} \|_{1,1})$ and $g(\mathbf{H}) = \infty 1_{(\| \mathbf{H} \|_* > K)} + \infty 1_{(\| \mathbf{H} \|_2 > 1)}$. 
Define $f_i(\boldsymbol{\Pi}) = - \tr \{ \hat{\mathbf{T}}_i \boldsymbol{\Pi} \} + \rho \| \boldsymbol{\Pi} \|_{1,1}$. 

First, we form the augmented Lagrangian for (\ref{eq2.11})
\begin{equation}
    \label{eqa.1}
    L(\boldsymbol{\Pi}, \mathbf{H}, \mathbf{G}) = \sum_{i=1}^{m} \frac{n_i}{N} f_i(\boldsymbol{\Pi}) + g(\mathbf{H}) + \langle \mathbf{G}, \hat{\boldsymbol{\Sigma}}^{1/2} \boldsymbol{\Pi} \hat{\boldsymbol{\Sigma}}^{1/2} - \mathbf{H} \rangle + \frac{\nu}{2} \|\hat{\boldsymbol{\Sigma}}^{1/2} \boldsymbol{\Pi} \hat{\boldsymbol{\Sigma}}^{1/2} - \mathbf{H}\|_{\mathrm{F}}^2, 
\end{equation}
where $\mathbf{G}$ is the dual variable, $\nu$ is the augmented Lagrangian parameter. 

The resulting ADMM algorithm is 
\begin{align*}
    \boldsymbol{\Pi}^{t+1} & := \argmin_{\boldsymbol{\Pi} \in \cM} \left (\sum_{i=1}^{m} \frac{n_i}{N} \{f_i(\boldsymbol{\Pi}) + \langle \mathbf{G}^t, \hat{\boldsymbol{\Sigma}}^{1/2} \boldsymbol{\Pi} \hat{\boldsymbol{\Sigma}}^{1/2} - \mathbf{H}^t \rangle + \frac{\nu}{2} \|\hat{\boldsymbol{\Sigma}}^{1/2} \boldsymbol{\Pi} \hat{\boldsymbol{\Sigma}}^{1/2} - \mathbf{H}^{t}\|_{\mathrm{F}}^2 \} \right ) \\
    \mathbf{H}^{t+1} & := \argmin_{\mathbf{H} \in \cM} \left ( g(\mathbf{H}) - \langle \mathbf{G}^t, \mathbf{H} \rangle + \frac{\nu}{2} \|\mathbf{H} - \hat{\boldsymbol{\Sigma}}^{1/2} \boldsymbol{\Pi}^{t+1} \hat{\boldsymbol{\Sigma}}^{1/2} \|_{\mathrm{F}}^2 \right ) \\
    \mathbf{G}^{t+1} & := \mathbf{G}^t + \nu (\hat{\boldsymbol{\Sigma}}^{1/2} \boldsymbol{\Pi}^{t+1} \hat{\boldsymbol{\Sigma}}^{1/2} - \mathbf{H}^{t+1} ). 
\end{align*}
With a scaled augmented Lagrangian parameter $\boldsymbol{\Gamma} = \mathbf{G} / \nu$, we can get the scaled form of this algorithm
\begin{align}
    \label{eq2.12a} \boldsymbol{\Pi}^{t+1} & := \argmin_{\boldsymbol{\Pi} \in \cM} \left ( \sum_{i=1}^{m} \frac{n_i}{N} \{ f_i(\boldsymbol{\Pi}) + \frac{\nu}{2} \|\hat{\boldsymbol{\Sigma}}^{1/2} \boldsymbol{\Pi} \hat{\boldsymbol{\Sigma}}^{1/2} - (\mathbf{H}^{t} - \boldsymbol{\Gamma}^{t})\|_{\mathrm{F}}^2 \} \right ) \\
    \label{eq2.12b} \mathbf{H}^{t+1} & := \argmin_{\mathbf{H} \in \cM} \left ( g(\mathbf{H}) + \frac{\nu}{2} \|\mathbf{H} - (\hat{\boldsymbol{\Sigma}}^{1/2} \boldsymbol{\Pi}^{t+1} \hat{\boldsymbol{\Sigma}}^{1/2} + \boldsymbol{\Gamma}^t) \|_{\mathrm{F}}^2 \right ) \\
    \label{eq2.12c} \boldsymbol{\Gamma}^{t+1} & := \boldsymbol{\Gamma}^t + \hat{\boldsymbol{\Sigma}}^{1/2} \boldsymbol{\Pi}^{t+1} \hat{\boldsymbol{\Sigma}}^{1/2} - \mathbf{H}^{t+1}.
\end{align}
The scaled version is simpler and easier to work with than the unscaled form. 

Update for $\boldsymbol{\Pi}$: (\ref{eq2.12a}) is composed of client-specific problems
\begin{equation}
    \label{eq2.13}
    \min_{\boldsymbol{\Pi} \in \cM} f_i(\boldsymbol{\Pi}) + \frac{\nu}{2} \|\hat{\boldsymbol{\Sigma}}^{1/2} \boldsymbol{\Pi} \hat{\boldsymbol{\Sigma}}^{1/2} - (\mathbf{H}^{t} - \boldsymbol{\Gamma}^{t})\|_{\mathrm{F}}^2. 
\end{equation}
To solve (\ref{eq2.12a}), we adopt the idea of Federated Primal-Dual method \citep{zhang2020fedpd}. 
We will solve the local problem first and average the solutions to get the closed-form update for $\boldsymbol{\Pi}$. 
We use vectorization operation to convert (\ref{eq2.13}) into a vector minimization form. 
Let $\boldsymbol{\pi} = \vct(\boldsymbol{\Pi})$, $\mathbf{h} = \vct(\mathbf{H})$, $\boldsymbol{\gamma} = \vct(\boldsymbol{\Gamma})$, $\boldsymbol{\tau}_i = \vct(\hat{\mathbf{T}}_i)$, and $\mathbf{C} = \hat{\boldsymbol{\Sigma}}^{1/2} \otimes \hat{\boldsymbol{\Sigma}}^{1/2}$. 
Use the identity $\vct(\mathbf{A}\mathbf{X}\mathbf{B}) = (\mathbf{B}^{\mathrm{T}} \otimes \mathbf{A}) \vct(\mathbf{X})$, (\ref{eq2.13}) turned into
\begin{equation}
    \label{eqa.2}
    \min_{\boldsymbol{\pi}} - \boldsymbol{\tau}_i^{\mathrm{T}} \boldsymbol{\pi} + \rho \| \boldsymbol{\pi} \|_1 + \frac{\nu}{2} \|\mathbf{C} \boldsymbol{\pi} - \mathbf{h} + \boldsymbol{\gamma}\|_2^2. 
\end{equation}

(\ref{eqa.2}) doesn't have a closed-form solution, but we can consider a linearized version of it by adding a quadratic term of $\boldsymbol{\pi} - \boldsymbol{\pi}^{t}$ as suggested by Fang et al. \cite{Fang2015GADMM}
\begin{equation}
    \label{eqa.3}
    \boldsymbol{\pi}_i^{t+} = \argmin_{\boldsymbol{\pi}} \left ( -\boldsymbol{\tau}_i^{\mathrm{T}} \boldsymbol{\pi} + \rho \| \boldsymbol{\pi} \|_1 + \nu \{ \boldsymbol{\pi} - \boldsymbol{\pi}^t \}^{\mathrm{T}} \mathbf{m}^t + \frac{\alpha_i}{2} \| \boldsymbol{\pi} - \boldsymbol{\pi}^t \|_2^2 \right ), 
\end{equation}
where $\mathbf{m}^t = \mathbf{C} (\mathbf{C} \boldsymbol{\pi}^t - \mathbf{h}^t + \boldsymbol{\gamma}^t)$. 
We choose $\alpha_i = 4 \nu \lambda^2_{\max} (\hat{\boldsymbol{\Sigma}}_i)$ to ensure the convergence of the algorithm. 

Convert (\ref{eqa.3}) into matrix optimization version: 
\begin{equation}
    \label{eqa.4}
    \boldsymbol{\Pi}_i^{t+} = \argmin_{\boldsymbol{\Pi} \in \cM} \rho \|\boldsymbol{\Pi}\|_{1,1} + \frac{\alpha_i}{2} \|\boldsymbol{\Pi} - [\boldsymbol{\Pi}^t + \frac{1}{\alpha_i} \hat{\mathbf{T}}_i - \frac{\nu}{\alpha_i} \mathbf{M}^t] \|_{\mathrm{F}}^2, 
\end{equation}
where $\mathbf{M}^t = \hat{\boldsymbol{\Sigma}} \boldsymbol{\Pi}^t \hat{\boldsymbol{\Sigma}} - \hat{\boldsymbol{\Sigma}}^{1/2} (\mathbf{H}^{t} - \boldsymbol{\Gamma}^{t}) \hat{\boldsymbol{\Sigma}}^{1/2}$. 
(\ref{eqa.4}) has the closed-form solution
\begin{equation}
    \label{eqa.5}
    \boldsymbol{\Pi}_i^{t+} = \mathrm{ST}[\boldsymbol{\Pi}^t + \frac{1}{\alpha_i} \hat{\mathbf{T}}_i - \frac{\nu}{\alpha_i} \mathbf{M}^t, \frac{\rho}{\alpha_i} ]. 
\end{equation}
Therefore, we can get the update of $\boldsymbol{\Pi}$: 
\begin{equation*}
    \boldsymbol{\Pi}^{t+1} = \sum_{i=1}^{m} \frac{n_i}{N} \boldsymbol{\Pi}_i^{t+}. 
\end{equation*}

\begin{remark}
We can write (\ref{eq2.13}) in vectorized form and then the problem is reduced to solving a Lasso regression problem
\begin{equation*}
    \min_{\boldsymbol{\pi}} \frac{\nu}{2} \|\mathbf{C} \boldsymbol{\pi} - \mathbf{d}_i^t\|_2^2 + \rho \|\boldsymbol{\pi}\|_1, 
\end{equation*}
where $\mathbf{d}_i^t = \vct(\mathbf{H}^t - \boldsymbol{\Gamma}^t + \frac{1}{\nu} \hat{\boldsymbol{\Sigma}}^{-1/2} \hat{\mathbf{T}}_i \hat{\boldsymbol{\Sigma}}^{-1/2})$. 
Although plenty of algorithms are proposed to solve Lasso regression problems, this transformation involves a $d^2$-dimensional Lasso problem in each iteration and increases the computation cost. 
\end{remark}

\begin{remark}
Observe that $$\|\boldsymbol{\Pi}^{t+1} - \boldsymbol{\Pi}^{t}\|_{\mathrm{F}} \leq \|\sum_{i=1}^{m} \frac{n_i}{N} \boldsymbol{\Pi}_i^{t+} - \boldsymbol{\Pi}^{t}\|_{\mathrm{F}} \leq \sum_{i=1}^{m} \frac{n_i}{N} \| \boldsymbol{\Pi}_i^{t+} - \boldsymbol{\Pi}^{t} \|_{\mathrm{F}}. $$
Theorem 6 in \cite{Fang2015GADMM} implies that $\| \boldsymbol{\Pi}_i^{t+} - \boldsymbol{\Pi}^{t} \|_{\mathrm{F}} \leq C_i / t$ for some constants $C_i$'s. 
Thus, we have $\|\boldsymbol{\Pi}^{t+1} - \boldsymbol{\Pi}^{t}\|_{\mathrm{F}} \leq C / t$ for some constant $C$. 
According to Lemma 4 in \cite{Fang2015GADMM}, if $\| \boldsymbol{\Pi}^{t+1} - \boldsymbol{\Pi}^{t} \|_{\mathrm{F}} = 0$, $\boldsymbol{\Pi}^{t+1}$ is a solution to (\ref{eq2.10}). 
A $\cO(1/t)$ convergence rate for our algorithm is thus established. 
\end{remark}

Update for $\mathbf{H}$: The update of $\mathbf{H}$ follows directly from Proposition 2 in \cite{Kean2018convex}. 

\begin{proposition}
\label{prop1}
\cite{Kean2018convex}
Let $\sum_{j=1}^d w_j \mathbf{u}_j \mathbf{u}_j^{\mathrm{T}}$ be the singular value decomposition of $\mathbf{W}$. 
Then the optimization problem $$\min_{\mathbf{H} \in \cM} \|\mathbf{H} - \mathbf{W}\|_{\mathrm{F}}^2 \text{ subject to } \|\mathbf{H}\|_* \leq K \text{ and } \|\mathbf{H}\|_2 \leq 1 $$ has solution $\mathbf{H}^* = \sum_{j=1}^d \min \{1, \max(w_j - \gamma^*, 0)\} \mathbf{u}_j \mathbf{u}_j^{\mathrm{T}}. $
$\gamma^*$ satisfies $$\gamma^* = \argmin_{\gamma > 0} \gamma \text{ subject to } \sum_{j=1}^d \min \{1, \max(w_j - \gamma, 0)\} \leq K. $$
\end{proposition}

Thus, let $\sum_{j=1}^d w_j \mathbf{u}_j \mathbf{u}_j^{\mathrm{T}}$ be the singular value decomposition of $\hat{\boldsymbol{\Sigma}}^{1/2} \boldsymbol{\Pi}^{t+1} \hat{\boldsymbol{\Sigma}}^{1/2} + \boldsymbol{\Gamma}^t$. 
By proposition \ref{prop1}, we have $$\mathbf{H}^{t+1} = \sum_{j=1}^d \min \{1, \max(w_j - \gamma^*, 0)\} \mathbf{u}_j \mathbf{u}_j^{\mathrm{T}}, $$where $$\gamma^* = \argmin_{\gamma > 0} \gamma \text{ subject to } \sum_{j=1}^d \min \{1, \max(w_j - \gamma, 0)\} \leq K. $$

\section{Proof of Theorem \ref{thm1}}

It is necessary to show that Lemma \ref{lem1} and Lemma \ref{lem2} hold to prove this theorem. 
In this section, we first show the proofs of Lemma \ref{lem1} and Lemma \ref{lem2}.
Then we are going to prove Theorem \ref{thm1}. 

\subsection{Proof of Lemma \ref{lem1}}

\begin{proof}
For simplify, we assume $\bE[\bx] = 0$ and $N = m n$ since each client's sample mean and sample size don't influence the theoretical results in this part. 
For pair $(j, k)$, we have 
$\hat{\boldsymbol{\Sigma}}_{jk} = \frac{1}{m} \sum_{i=1}^{m} \frac{1}{n} \sum_{l_i=1}^{n} \bx_{l_i,j}^{(i)} \bx_{l_i,k}^{(i)} = \frac{1}{m n} \sum_{(i, l_i)=(1,1)}^{(m,n)} \bx_{l_i,j}^{(i)} \bx_{l_i,k}^{(i)}$
where $\hat{\boldsymbol{\Sigma}}_{jk}$ is the $(j,k)$th element of matrix $\hat{\boldsymbol{\Sigma}}$, and $\bx_{l_i,j}^{(i)}$ is the $j$th element of $\bx_{l_i}^{(i)}$. 
Then Lemma 1 in \cite{ravikumar2011high} implies the results of Lemma \ref{lem1} here. 
\end{proof}

\subsection{Proof of Lemma \ref{lem2}}

\begin{proof}
Recall that $\bx^{(i)}_{(k)^*}$ are the concomitants of the order statistics $y^{(i)}_{(k)}$ and $m^{i}(y)$ is the inverse regression curve at client $i$. 
Define the model error of the inverse regression by $\boldsymbol{\epsilon}^{(i)} = \bx^{(i)} - m^{i}(y^{(i)})$. 
Clearly, $\eps{i}{k} = \xc{i}{k} - \my{i}{k}$ are also the concomitants of $y^{(i)}_{(k)}$. 
(2.1) and (2.4) in \cite{Yang1977concomitant} show that the concomitants are conditionally independent given order statistics $y^{(i)}_{(k)}$ and $\bE[\xc{i}{k}] = \bE[\my{i}{k}]$. 
Then, $\eps{i}{k}$ are zero-mean random vectors and conditionally independent given $y^{(i)}_{(k)}$. 

First, we will introduce sub-Gaussian and sub-exponential distributions. 
For a random variable $z \in \bR$, define its $\psi_1$-norm and $\psi_2$-norm as $\|z\|_{\psi_1} = \sup_{p \geq 1} p^{-1} (\bE \lvert z \rvert^p)^{1/p}$ and $\|z\|_{\psi_2} = \sup_{p \geq 1} p^{-1/2} (\bE \lvert z \rvert^p)^{1/p}$. 
Sub-Gaussian random variable is defined to have a $\psi_2$-norm less than a constant and sub-exponential random variable is defined to have a $\psi_1$-norm less than a constant. 
The following proposition states the connection between sub-Gaussian and sub-exponential random variables. 
The proof of it is a direct consequence of H\"older's inequality and the definitions of sub-Gaussian and sub-exponential random variables. 
\begin{proposition}
\label{prop2}
Suppose $X$ and $Y$ are sub-Gaussian random variables, then $X Y$ is sub-exponential and $\|X Y\|_{\psi_1} \leq 2 \|X\|_{\psi_2} \|Y\|_{\psi_2}$. 
\end{proposition}

Denote the $j$-th element of $\bx^{(i)}$ by $\bx^{(i)}_j$, the same for $\boldsymbol{\epsilon}^{(i)}_j$ and $m^{i}_j(y)$. 
Clearly, under Condition (C2), $\bx^{(i)}_j$ is a sub-Gaussian random variable and its $\psi_2$-norm $\|\bx^{(i)}_j\|_{\psi_2} \leq C (\boldsymbol{\Sigma}_{j j})^{1/2}$. 
Proposition 3 in \cite{Kean2018convex} states that $m^i_j(y^{(i)})$ and $\boldsymbol{\epsilon}^{(i)}_j$ are also sub-Gaussian random variables with norm $\|\bx^{(i)}_j\|_{\psi_2}$ and $2 \|\bx^{(i)}_j\|_{\psi_2}$, respectively. 

Now, we have 
\begin{align*}
    \hat{\mathbf{Q}} & = \frac{1}{N} \sum_{i=1}^{m} n_i \hat{\mathbf{Q}}_i \\
    & = \frac{1}{N} \sum_{i=1}^{m} \sum_{k=1}^{\lfloor n_i/2 \rfloor} \{\bx^{(i)}_{(2k)^*} - \bx^{(i)}_{(2k-1)^*}\}\{\bx^{(i)}_{(2k)^*} - \bx^{(i)}_{(2k-1)^*}\}^{\mathrm{T}} \\
    & = \frac{1}{N} \sum_{i=1}^{m} \sum_{k=1}^{\lfloor n_i/2 \rfloor} \{ \my{i}{2k} - \my{i}{2k-1} \} \{ \my{i}{2k} - \my{i}{2k-1} \}^{\mathrm{T}} \\
    & + \frac{1}{N} \sum_{i=1}^{m} \sum_{k=1}^{\lfloor n_i/2 \rfloor} \{ \eps{i}{2k} - \eps{i}{2k-1} \} \{ \my{i}{2k} - \my{i}{2k-1} \}^{\mathrm{T}} \\
    & + \frac{1}{N} \sum_{i=1}^{m} \sum_{k=1}^{\lfloor n_i/2 \rfloor} \{ \my{i}{2k} - \my{i}{2k-1} \} \{ \eps{i}{2k} - \eps{i}{2k-1} \}^{\mathrm{T}} \\
    & + \frac{1}{N} \sum_{i=1}^{m} \sum_{k=1}^{\lfloor n_i/2 \rfloor} \{ \eps{i}{2k} - \eps{i}{2k-1} \} \{ \eps{i}{2k} - \eps{i}{2k-1} \}^{\mathrm{T}} \\
    & = M_1 + M_2 + M_3 + M_4. \\ 
\end{align*}

It is easy to see that 
\begin{equation}
\label{eq.p1}
   \|\hat{\mathbf{Q}} - \bar{\mathbf{Q}}\|_{\max} \leq \|\mathbf{M}_1\|_{\max} + \|\mathbf{M}_2\|_{\max} + \|\mathbf{M}_3\|_{\max} + \|\mathbf{M}_4 - \mathbf{Q}^*\|_{\max} + \|\mathbf{Q}^* - \bar{\mathbf{Q}}\|_{\max}, 
\end{equation}
where $\mathbf{Q}^* = \sum_{i=1}^{m} \frac{n_i}{N} \mathbf{Q}_i$. 
Then we only need to derive the convergence rate of the foregoing five terms on the right side of (\ref{eq.p1}). 

For the $(j,l)$th element of $\mathbf{M}_1$, we have 
\begin{align*}
    \lvert (\mathbf{M}_1)_{j l} \rvert & = \lvert \frac{1}{N} \sum_{i=1}^{m} \sum_{k=1}^{\lfloor n_i/2 \rfloor} \{ \m{i}{2k}{j} - \m{i}{2k-1}{j} \} \{ \m{i}{2k}{l} - \m{i}{2k-1}{l} \} \rvert \\
    & \leq \frac{1}{N} \sum_{i=1}^{m} [ \sup_{\cU(B)} \sum_{k=2}^{n_i} \|\my{i}{k} - \my{i}{k-1} \|_{\infty}]^2 \\
    & \leq \frac{m \delta^2}{N}. 
\end{align*}
By Condition (C4), there exists some constant $\delta > 0$, we have $\sup_{\cU(B)} \sum_{k=2}^{n} \|\my{i}{k} - \my{i}{k-1} \|_{\infty} \leq \delta$, which implies the last inequality above. 
Thus, we have $\|\mathbf{M}_1\|_{\max} \leq \frac{m \delta^2}{N} \leq \frac{C \delta^2}{N^{1 - \eta}}$. 

Similarly, for the $(j, l)$th element of $\mathbf{M}_2$, we have 
\begin{align*}
    \lvert (\mathbf{M}_2)_{j l}\rvert & = \lvert \frac{1}{N} \sum_{i=1}^{m} \sum_{k=1}^{\lfloor n_i/2 \rfloor} \{\ep{i}{2k}{j} - \ep{i}{2k-1}{j}\} \{\m{i}{2k}{j} - \m{i}{2k-1}{j}\} \rvert \\ 
    & \leq \frac{1}{N} \sum_{i=1}^{m} 2 \max_{k,j} \lvert \boldsymbol{\epsilon}^{(i)}_{k,j} \rvert [\sup_{\cU(B)} \sum_{k=2}^{n_i} \|\my{i}{k} - \my{i}{k-1} \|_{\infty}] \\
    & \leq \sum_{i=1}^{m} \frac{2 \delta}{N} \max_{k,j} \lvert \boldsymbol{\epsilon}^{(i)}_{k,j} \rvert. \\
\end{align*}
Then, following directly from the equivalence of sub-Gaussian properties in Lemma 5.5 in \cite{vershynin2011introduction}, for some constant $C$, we have $$\mathrm{Pr} \left (\max_{k,j} \lvert \boldsymbol{\epsilon}^{(i)}_{k,j} \rvert \geq C (\frac{N^{2 \eta}}{m^{2}}\log d)^{1/2}\right ) \leq \exp (-C' \frac{N^{2 \eta}}{m^{2}} \log d), $$ with $C' \frac{N^{2 \eta}}{m^{2}} > 2$. 
Thus, with probability being at least $1 - \exp(-C' \frac{N^{2 \eta}}{m^{2}} \log d)$, we have $\lvert (\mathbf{M}_2)_{j l} \rvert \leq C\frac{(\log d)^{1/2}}{N^{1 - \eta}}$. 
Taking the union bound, we have 
\begin{align*}
    \mathrm{Pr} \left (\|\mathbf{M}_2\|_{\max} \geq C\frac{(\log d)^{1/2}}{N^{1 - \eta}} \right ) & \leq \sum_{j,l} \mathrm{Pr} \left (\lvert (\mathbf{M}_2)_{j l} \rvert \geq C\frac{(\log d)^{1/2}}{N^{1 - \eta}} \right ) \\
    & \leq d^2 \mathrm{Pr} \left (\lvert (\mathbf{M}_2)_{j l} \rvert \geq C\frac{(\log d)^{1/2}}{N^{1 - \eta}} \right ) \\
    & \leq \exp(-C' \frac{N^{2 \eta}}{m^{2}} \log d + 2 \log d) \\
    & \leq \exp(-C'' \log d). 
\end{align*}
$\mathbf{M}_3$ can also be upper bounded similarly. 

For the $(j,l)$th element of $\mathbf{M}_4 - \mathbf{Q}^*$, we have
\begin{align*}
    (\mathbf{M}_4)_{j l} - \mathbf{Q}^{*}_{j l} & = \frac{1}{N} \sum_{i=1}^{m} \sum_{k=1}^{\lfloor n_i/2 \rfloor} \{ \ep{i}{2k}{j} - \ep{i}{2k-1}{j} \} \{ \ep{i}{2k}{l} - \ep{i}{2k-1}{l} \} - \mathbf{Q}^{*}_{j l} \\
    & = \frac{1}{N} \sum_{i=1}^{m} [ \sum_{k=1}^{n_i} (\boldsymbol{\epsilon}^{(i)}_{k, j} \boldsymbol{\epsilon}^{(i)}_{k, l} - \mathbf{Q}_{i, j l}) - \sum_{k=1}^{\lfloor n_i/2 \rfloor} \ep{i}{2k}{j} \ep{i}{2k-1}{l} \\ 
    & - \sum_{k=1}^{\lfloor n_i/2 \rfloor} \ep{i}{2k}{l} \ep{i}{2k-1}{j}] \\
    & = \mathbf{J}_1 - \mathbf{J}_2 - \mathbf{J}_3. 
\end{align*}
It requires to show that $\mathbf{J}_1$, $\mathbf{J}_2$ and $\mathbf{J}_3$ is upper bounded by $C(\log d/ N)^{1/2}$. 

Since $\bE[\boldsymbol{\epsilon}^{(i)} \boldsymbol{\epsilon}^{(i)T}] = \bE[(\bx^{(i)} - m^{i}(y^{(i)}))(\bx^{(i)} - m^{i}(y^{(i)}))^{\mathrm{T}}] = \bE[\bx^{(i)} \bx^{(i)T}] - \bE[\bE(\bx^{(i)} \vert y^{(i)})\bE(\bx^{(i)} \vert y^{(i)})^{\mathrm{T}}] = E[\cov(\bx^{(i)} \vert y^{(i)})] = \mathbf{Q}_i$, combined with Proposition \ref{prop2} and the centering lemma in \cite{vershynin2018HDP}, $\boldsymbol{\epsilon}^{(i)}_{k, j}\boldsymbol{\epsilon}^{(i)}_{k,l} - \mathbf{Q}_{i, j l}$ is sub-exponential with $\psi_1$-norm bounded by $c \| \bx\|_{\psi_2}^2$, where $c > 0$ is a constant. 
Thus, following directly from the Bernstein-type inequality in Proposition 5.16 in \cite{vershynin2011introduction}, for some sufficiently large constant $C$, we obtain $$\mathrm{Pr} \left (\lvert \frac{1}{N} \sum_{i=1}^{m} \sum_{k=1}^{n_i} (\boldsymbol{\epsilon}^{(i)}_{k, j} \boldsymbol{\epsilon}^{(i)}_{k, l} - \mathbf{Q}_{i, j l})\rvert \geq C (\frac{\log d}{N})^{1/2} \right ) \leq \exp(-C' \log d). $$
Also, by taking the union bound, we have $$\|\mathbf{J}_1\|_{\max} \leq C (\frac{\log d}{N})^{1/2}, $$ with probability greater than $1 - \exp(-C'' \log d)$. 

Recall that $\eps{i}{2k}$ and $\eps{i}{2k-1}$ are conditionally independent sub-Gaussian random vectors with mean zero given the order statistics $y^{(i)}_{(1)}, \dots, y^{(i)}_{(n_i)}$. 
Thus, $\ep{i}{2k}{j} \ep{i}{2k-1}{l}$ is sub-exponential with mean zero and have a $\psi_1$-norm less than $c \|\bx\|_{\psi_2}^2$, conditionally given the order statistics. 
Similar to above, by the Berstein-type inequality and taking the union bound, we have $$\mathrm{Pr} \left (\|\mathbf{J}_2\|_{\max} \geq C (\frac{\log d}{N})^{1/2} \mid y^{(i)}_{(1)}, \dots, y^{(i)}_{(n_i)}, i \in [m] \right ) \leq \exp(- C'' \log d). $$
Taking expectation on the order statistics, we have
\begin{align*}
    \mathrm{Pr} \left (\|\mathbf{J}_2\|_{\max} \geq C (\frac{\log d}{N})^{1/2} \right ) & = \bE \left [ \mathrm{Pr} \left (\|\mathbf{J}_2\|_{\max} \geq C (\frac{\log d}{N})^{1/2} \mid y^{(i)}_{(1)}, \dots, y^{(i)}_{(n_i)}, i \in [m] \right ) \right ] \\
    & \leq \exp(-C'' \log d). 
\end{align*}
$\mathbf{J}_3$ can also be upper bounded in the same manner. 
Thus, with probability being at least $1 - \exp(- C' \log d)$, $\|\mathbf{M}_4 - \mathbf{Q}^{*}\|_{\max}$ is upper bounded by $C (\frac{\log d}{N})^{1/2}$. 

For the $(j, l)$th element of $\mathbf{Q}^* - \bar{\mathbf{Q}}$, we have
\begin{align*}
    \lvert \mathbf{Q}^{*}_{j l} - \bar{\mathbf{Q}}_{j l} \rvert & = \sum_{i=1}^{m} \lvert \frac{n_i}{N} - \omega_i \rvert \lvert Q_{i, j l} \rvert \\ 
    & \leq C \sum_{i=1}^{m} \lvert \frac{n_i}{N} - \omega_i \rvert. \\
\end{align*}
This inequality follows from Condition (C3) and (C6). 
By Lemma C.2 in \cite{Agrawal2017rl}, we have an inequality against multinomial random vector $$\mathrm{Pr} \left (\sum_{i=1}^{m} \lvert \frac{n_i}{N} - \omega_i \rvert \geq C (\frac{\log d}{N})^{1/2} \right ) \leq \exp (-C' \log d). $$
Taking the union bound, for some constants $C, C^{\prime}, C^{\prime\prime}$, we have
\begin{align*}
    \mathrm{Pr} \left (\| \mathbf{Q}^* - \bar{\mathbf{Q}} \|_{\max} \geq C (\frac{\log d}{N})^{1/2} \right ) & \leq \sum_{j,l} \mathrm{Pr} \left ( \lvert \mathbf{Q}^{*}_{j l} - \bar{\mathbf{Q}}_{j l} \rvert \geq C (\frac{\log d}{N})^{1/2} \right ) \\ 
    & \leq d^2 \mathrm{Pr} \left ( \lvert \mathbf{Q}^{*}_{j l} - \bar{\mathbf{Q}}_{j l} \rvert \geq C (\frac{\log d}{N})^{1/2} \right ) \\
    & \leq d^2 \mathrm{Pr} \left ( \sum_{i=1}^{m} \lvert \frac{n_i}{N} - \omega_i \rvert \geq C^{\prime} (\frac{\log d}{N})^{1/2} \right ) \\
    & \leq \exp (-C^{\prime\prime} \log d). \\
\end{align*}

Together with the upper bounds of $\mathbf{M}_1$, $\mathbf{M}_2$, $\mathbf{M}_3$ and $\mathbf{M}_4 - \mathbf{Q}^*$, we have $$\|\hat{\mathbf{Q}} - \bar{\mathbf{Q}}\|_{\max} \leq C (\frac{\log d}{N})^{1/2} + C^{\prime} \frac{(\log d)^{1/2}}{N^{1 - \eta}}, $$ with probability being at least $1 - \exp(-C_1 \log d)$ for some constant $C_1$. 
\end{proof}

\subsection{Proof of Theorem \ref{thm1}}

\begin{proof}

Before establishing the upper bound on the distance between true and estimated subspaces, we first consider the statistical error of the estimated projection matrix $\hat{\Pi}$. 

Proposition 4 in \cite{Kean2018convex} states that $\mathbf{V}$ is the solution of (\ref{eq2.5c}) if and only if $\mathbf{T}$ can be written as $\boldsymbol{\Sigma} \mathbf{V} \boldsymbol{\Lambda} \mathbf{V}^{\mathrm{T}} \boldsymbol{\Sigma}$, where $\mathbf{V}^{\mathrm{T}} \boldsymbol{\Sigma} \mathbf{V} = \mathbf{I}_{K}$. 
Let $\tilde{\mathbf{T}} = \hat{\boldsymbol{\Sigma}} \mathbf{V} \boldsymbol{\Lambda} \mathbf{V}^{\mathrm{T}} \hat{\boldsymbol{\Sigma}}$, $\tilde{\mathbf{V}} = \mathbf{V} (\mathbf{V}^{\mathrm{T}} \hat{\boldsymbol{\Sigma}} \mathbf{V})^{-1/2}$, $\tilde{\boldsymbol{\Pi}} = \tilde{\mathbf{V}} \tilde{\mathbf{V}}^{\mathrm{T}}$ and $\tilde{\boldsymbol{\Lambda}} = (\mathbf{V}^{\mathrm{T}} \hat{\boldsymbol{\Sigma}} \mathbf{V})^{1/2} \boldsymbol{\Lambda} (\mathbf{V}^{\mathrm{T}} \hat{\boldsymbol{\Sigma}} \mathbf{V})^{1/2}$. 
$\tilde{\boldsymbol{\Pi}}$ can be viewed as the bridge between $\hat{\boldsymbol{\Pi}}$ and $\boldsymbol{\Pi}$. 
Thus, we obtain 
\begin{equation*}
    \| \hat{\boldsymbol{\Pi}} - \boldsymbol{\Pi} \|_{\mathrm{F}} \leq \| \tilde{\boldsymbol{\Pi}} - \boldsymbol{\Pi} \|_{\mathrm{F}} + \| \hat{\boldsymbol{\Pi}} - \tilde{\boldsymbol{\Pi}} \|_{\mathrm{F}}. 
\end{equation*}

For the term $\| \tilde{\boldsymbol{\Pi}} - \boldsymbol{\Pi} \|_{\mathrm{F}}$, Lemma 2 in \cite{Kean2018convex} establishes concentration that given constants $C_1$ and $C_2$, 
\begin{equation}
\label{b3.0}
    \| \tilde{\boldsymbol{\Pi}} - \boldsymbol{\Pi} \|_{\mathrm{F}} \leq C_1 K (s / N)^{1/2} \leq C_2 s \rho. 
\end{equation}
with probability being at least $1 - \exp(-C' s)$. 
The second inequality holds by Condition (C5) and the assumption that $\rho \geq C (\log d / N)^{1/2}$ for some constant $C$. 

Let $\boldsymbol{\Delta} = \hat{\boldsymbol{\Pi}} - \tilde{\boldsymbol{\Pi}}$. 
Following Gao et al. \cite{Gao2017scca}, we first derive the upper bound and lower bound for $\| \hat{\boldsymbol{\Sigma}}^{1/2} \boldsymbol{\Delta} \hat{\boldsymbol{\Sigma}}^{1/2} \|_{\mathrm{F}}$. 
Then we can upper bound $\boldsymbol{\Delta}$ with those results. 

Upper bound for $\| \hat{\boldsymbol{\Sigma}}^{1/2} \boldsymbol{\Delta} \hat{\boldsymbol{\Sigma}}^{1/2} \|_{\mathrm{F}}$: 
First, we want to show that $\tilde{\boldsymbol{\Pi}}$ is a feasible solution of (\ref{eq2.10}). 
We have 
\begin{align*}
    \| \hat{\boldsymbol{\Sigma}}^{1/2} \tilde{\boldsymbol{\Pi}} \hat{\boldsymbol{\Sigma}}^{1/2} \|_2 & \leq \|\hat{\boldsymbol{\Sigma}}^{1/2} \tilde{\mathbf{V}} \|_2^2  = \|\tilde{\mathbf{V}}^{\mathrm{T}} \hat{\boldsymbol{\Sigma}} \tilde{\mathbf{V}} \|_2 \\
    & = \|\left(\mathbf{V}^{\mathrm{T}} \hat{\boldsymbol{\Sigma}} \mathbf{V}\right)^{-1 / 2} \mathbf{V}^{\mathrm{T}} \hat{\boldsymbol{\Sigma}} \mathbf{V}\left(\mathbf{V}^{\mathrm{T}} \hat{\boldsymbol{\Sigma}} \mathbf{V}\right)^{-1 / 2}\|_2 \\
    & = \|\mathbf{I}_K\|_2 = 1, 
\end{align*}
and
\begin{align*}
    \| \hat{\boldsymbol{\Sigma}}^{1/2} \tilde{\boldsymbol{\Pi}} \hat{\boldsymbol{\Sigma}}^{1/2} \|_* & = \tr\{\hat{\boldsymbol{\Sigma}}^{1/2} \tilde{\boldsymbol{\Pi}} \hat{\boldsymbol{\Sigma}}^{1/2}\} \\
    & = \tr\{\tilde{\mathbf{V}}^{\mathrm{T}} \hat{\boldsymbol{\Sigma}} \tilde{\mathbf{V}}\} \\
    & = K. 
\end{align*}
Since $\hat{\boldsymbol{\Sigma}}$ is the optimal solution of (\ref{eq2.10}), we have 
\begin{equation*}
    -\langle \hat{\mathbf{T}}, \hat{\boldsymbol{\Pi}} \rangle+\rho\|\hat{\boldsymbol{\Pi}} \|_{1,1} \leq - \langle \hat{\mathbf{T}}, \tilde{\boldsymbol{\Pi}} \rangle+\rho\|\tilde{\boldsymbol{\Pi}}\|_{1,1}. 
\end{equation*}
Rearrange the variables and use the H\"older's inequality we have
\begin{align*}
    \rho\|\tilde{\boldsymbol{\Pi}}\|_{1,1} - \rho\|\tilde{\boldsymbol{\Pi}} + \boldsymbol{\Delta}\|_{1,1} & \geq - \langle \tilde{\mathbf{T}}, \boldsymbol{\Delta} \rangle - \langle \hat{\mathbf{T}} - \tilde{\mathbf{T}}, \boldsymbol{\Delta} \rangle \\
    & \geq - \langle \tilde{\mathbf{T}}, \boldsymbol{\Delta} \rangle - \|\hat{\mathbf{T}} - \tilde{\mathbf{T}} \|_{\max} \|\boldsymbol{\Delta}\|_{1,1}. 
\end{align*}
When $\|\hat{\mathbf{T}} - \tilde{\mathbf{T}} \|_{\max} \leq \frac{\rho}{2}$, we have
\begin{equation}
    \label{b3.1}
    \rho\|\tilde{\boldsymbol{\Pi}}\|_{1,1} - \rho\|\tilde{\boldsymbol{\Pi}} + \boldsymbol{\Delta}\|_{1,1} + \frac{\rho}{2} \|\boldsymbol{\Delta}\|_{1,1} \geq - \langle \tilde{\mathbf{T}}, \boldsymbol{\Delta} \rangle. 
\end{equation}

Let $\cS = \mathrm{supp}(\boldsymbol{\Pi})$, we have $\lvert \cS \rvert \leq s^2$. 
Let $\cS^c$ be the complementary set of $\cS$. 
By the definition of $\tilde{\boldsymbol{\Pi}}$ and $\tilde{\mathbf{V}}$, we have $\mathrm{supp}(\tilde{\boldsymbol{\Pi}}) = \mathrm{supp}(\boldsymbol{\Pi})$. 
Reconsider the LHS of (\ref{b3.1}), 
\begin{align*}
    \rho\|\tilde{\boldsymbol{\Pi}} \|_{1,1} - \rho\|\tilde{\boldsymbol{\Pi}} + \boldsymbol{\Delta}\|_{1,1} + \frac{\rho}{2} \|\boldsymbol{\Delta}\|_{1,1} & = \rho\|\tilde{\boldsymbol{\Pi}}_{\cS} \|_{1,1} - \rho\|\tilde{\boldsymbol{\Pi}}_{\cS} + \boldsymbol{\Delta}_{\cS}\|_{1,1} - \rho \|\boldsymbol{\Delta}_{\cS^c}\|_{1,1} + \frac{\rho}{2} \|\boldsymbol{\Delta}\|_{1,1} \\
    & \leq \rho \|\boldsymbol{\Delta}_{\cS}\|_{1,1} - \rho \|\boldsymbol{\Delta}_{\cS^c}\|_{1,1} + \frac{\rho}{2} \|\boldsymbol{\Delta}_{\cS} + \boldsymbol{\Delta}_{\cS^c} \|_{1,1} \\
    & = \frac{3\rho}{2} \|\boldsymbol{\Delta}_{\cS}\|_{1,1} - \frac{\rho}{2} \|\boldsymbol{\Delta}_{\cS^c}\|_{1,1}. 
\end{align*}
Thus, we have
\begin{equation}
\label{b3.2}
    - \langle \tilde{\mathbf{T}}, \boldsymbol{\Delta} \rangle \leq \frac{3\rho}{2} \|\boldsymbol{\Delta}_{\cS}\|_{1,1} - \frac{\rho}{2} \|\boldsymbol{\Delta}_{\cS^c}\|_{1,1}. 
\end{equation}
Directly from Lemma 3 in \cite{Kean2018convex}, we obtain 
\begin{equation}
\label{b3.3}
\begin{aligned}
- \langle \tilde{\mathbf{T}}, \boldsymbol{\Delta} \rangle &= \langle\hat{\boldsymbol{\Sigma}}^{1/2} \mathbf{V} \boldsymbol{\Lambda} \mathbf{V}^{\mathrm{T}} \hat{\boldsymbol{\Sigma}}^{1/2}, \hat{\boldsymbol{\Sigma}}^{1/2}(\tilde{\boldsymbol{\Pi}}-\hat{\boldsymbol{\Pi}}) \hat{\boldsymbol{\Sigma}}^{1/2} \rangle \\
&= \langle\hat{\boldsymbol{\Sigma}}^{1/2} \tilde{\mathbf{V}} \tilde{\boldsymbol{\Lambda}} \tilde{\mathbf{V}}^{\mathrm{T}} \hat{\boldsymbol{\Sigma}}^{1/2}, \hat{\boldsymbol{\Sigma}}^{1/2}(\tilde{\boldsymbol{\Pi}}-\hat{\boldsymbol{\Pi}}) \hat{\boldsymbol{\Sigma}}^{1/2} \rangle \\
& \geq \frac{\lambda_{K}}{2}\|\hat{\boldsymbol{\Sigma}}^{1/2}(\tilde{\boldsymbol{\Pi}}-\hat{\boldsymbol{\Pi}}) \hat{\boldsymbol{\Sigma}}^{1/2} \|_{F}^{2}-\|\tilde{\boldsymbol{\Lambda}}-\boldsymbol{\Lambda}\|_{F} \|\hat{\boldsymbol{\Sigma}}^{1/2}(\tilde{\boldsymbol{\Pi}}-\hat{\boldsymbol{\Pi}}) \hat{\boldsymbol{\Sigma}}^{1/2} \|_{F}. 
\end{aligned}
\end{equation}
Let $\delta = \|\tilde{\boldsymbol{\Lambda}}-\boldsymbol{\Lambda}\|_{F}$. 
Combining (\ref{b3.2}) and (\ref{b3.3}), we have
\begin{equation}
\label{b3.4}
\begin{aligned}
    \lambda_{K}\|\hat{\boldsymbol{\Sigma}}^{1/2} \boldsymbol{\Delta} \hat{\boldsymbol{\Sigma}}^{1/2}\|_{\mathrm{F}}^{2} - 2 \delta \|\hat{\boldsymbol{\Sigma}}^{1/2} \boldsymbol{\Delta} \hat{\boldsymbol{\Sigma}}^{1/2}\|_{\mathrm{F}} & \leq 3 \rho\|\boldsymbol{\Delta}_{\mathcal{S}}\|_{1,1} - \rho\|\boldsymbol{\Delta}_{\cS^c}\|_{1,1} \\ 
    & \leq 3 \rho\|\boldsymbol{\Delta}_{\mathcal{S}}\|_{1,1}, 
\end{aligned}
\end{equation}
which implies 
\begin{equation}
\label{b3.5}
    \|\hat{\boldsymbol{\Sigma}}^{1/2} \boldsymbol{\Delta} \hat{\boldsymbol{\Sigma}}^{1/2}\|_{\mathrm{F}}^{2} \leq \frac{4 \delta^{2}}{\lambda_{K}^{2}}+\frac{6 \rho}{\lambda_{K}}\|\boldsymbol{\Delta}_{\mathcal{S}}\|_{1,1}. 
\end{equation}

Lower bound for $\| \hat{\boldsymbol{\Sigma}}^{1/2} \boldsymbol{\Delta} \hat{\boldsymbol{\Sigma}}^{1/2} \|_{\mathrm{F}}$: 
By (\ref{b3.4}), we have 
\begin{equation}
\label{b3.6}
    2 \delta \|\hat{\boldsymbol{\Sigma}}^{1/2} \boldsymbol{\Delta} \hat{\boldsymbol{\Sigma}}^{1/2}\|_{\mathrm{F}} + 3 \rho\|\boldsymbol{\Delta}_{\mathcal{S}}\|_{1,1} - \rho\|\boldsymbol{\Delta}_{\cS^c}\|_{1,1} \geq 0. 
\end{equation}
Using the fact that $a^2 + b^2 \geq 2 a b$, we have 
\begin{equation}
\label{b3.7}
    \frac{\delta^2}{\lambda_K} + \lambda_K \|\hat{\boldsymbol{\Sigma}}^{1/2} \boldsymbol{\Delta} \hat{\boldsymbol{\Sigma}}^{1/2}\|_{\mathrm{F}}^2 \geq 2 \delta \|\hat{\boldsymbol{\Sigma}}^{1/2} \boldsymbol{\Delta} \hat{\boldsymbol{\Sigma}}^{1/2}\|_{\mathrm{F}}. 
\end{equation}
Combining (\ref{b3.5}), (\ref{b3.6}) and (\ref{b3.7}), we obtain
\begin{equation}
\label{b3.8}
    \frac{5 \delta^2}{\lambda_K} + 9 \rho\|\boldsymbol{\Delta}_{\mathcal{S}}\|_{1,1} - \rho\|\boldsymbol{\Delta}_{\cS^c}\|_{1,1} \geq 0. 
\end{equation}
(\ref{b3.8}) shows that $\boldsymbol{\Delta}$ lies in a restricted set. 
We further partition the set $\cS^c$ into $J$ subsets. 
$\cS^c_1$ contains the indices corresponding to the largest $l$ entries of $\lvert \boldsymbol{\Delta} \rvert$, $\cS^c_2$ contains the indices corresponding to the second largest $l$ entries of $\lvert \boldsymbol{\Delta} \rvert$, and so forth, with $\lvert \cS^c_J \rvert \leq l$. 
By the triangle inequality, we have
\begin{equation}
\label{b3.10}
\begin{aligned}
\|\hat{\boldsymbol{\Sigma}}^{1 / 2} \boldsymbol{\Delta} \hat{\boldsymbol{\Sigma}}^{1 / 2}\|_{\mathrm{F}} & \geq\|\hat{\boldsymbol{\Sigma}}^{1 / 2} \boldsymbol{\Delta}_{\mathcal{S} \cup \mathcal{S}_{1}^{c}} \hat{\boldsymbol{\Sigma}}^{1 / 2}\|_{\mathrm{F}}-\sum_{j=2}^{J}\|\hat{\boldsymbol{\Sigma}}^{1 / 2} \boldsymbol{\Delta}_{\mathcal{S}_{j}^{c}} \hat{\boldsymbol{\Sigma}}^{1 / 2}\|_{\mathrm{F}} \\
& \geq \lambda_{\min }(\hat{\boldsymbol{\Sigma}}, s+l)\|\boldsymbol{\Delta}_{\mathcal{S} \cup \mathcal{S}_{1}^{c}}\|_{\mathrm{F}}-\lambda_{\max }(\hat{\boldsymbol{\Sigma}}, l) \sum_{j=2}^{J}\|\boldsymbol{\Delta}_{\mathcal{S}_{j}^{c}}\|_{\mathrm{F}}. 
\end{aligned}
\end{equation}
The definitions of $\lambda_{\max}(\hat{\boldsymbol{\Sigma}}, s)$ and $\lambda_{\max}(\hat{\boldsymbol{\Sigma}}, s)$ are deferred to Appendix \ref{defs}. 
The second inequality follows directly from the definitions. 
For $\|\boldsymbol{\Delta}_{\mathcal{S}_{j}^{c}}\|_{\mathrm{F}}$, we have
\begin{equation*}
    \sum_{j=2}^{J} \|\boldsymbol{\Delta}_{\mathcal{S}_{j}^{c}}\|_{\mathrm{F}} \leq l^{1/2} \sum_{j=2}^{J} \|\boldsymbol{\Delta}_{\mathcal{S}_{j}^{c}}\|_{\max} \leq l^{-1/2} \sum_{j=2}^{J} \|\boldsymbol{\Delta}_{\mathcal{S}_{j-1}^{c}}\|_{1, 1} \leq l^{-1/2} \|\boldsymbol{\Delta}_{\cS^c}\|_{1, 1}, 
\end{equation*}
which is combined with (\ref{b3.8}), to obtain
\begin{equation}
\label{b3.9}
\begin{aligned}
    \sum_{j=2}^{J} \|\boldsymbol{\Delta}_{\mathcal{S}_{j}^{c}}\|_{\mathrm{F}} & \leq \frac{5 \delta^2}{\rho \lambda_K l^{1/2}} + 9 l^{-1/2} \|\boldsymbol{\Delta}_{\mathcal{S}}\|_{1,1} \\
    & \leq \frac{5 \delta^2}{\rho \lambda_K l^{1/2}} + 9 s l^{-1/2} \|\boldsymbol{\Delta}_{\mathcal{S}}\|_{\mathrm{F}}, 
\end{aligned}
\end{equation}
where the second inequality follows from the Cauchy-Schwarz inequality. 

Substituting (\ref{b3.9}) into (\ref{b3.10}), we have 
\begin{equation}
\label{b3.11}
\begin{aligned}
    \|\hat{\boldsymbol{\Sigma}}^{1 / 2} \boldsymbol{\Delta} \hat{\boldsymbol{\Sigma}}^{1 / 2}\|_{\mathrm{F}} & \geq \lambda_{\min }(\hat{\boldsymbol{\Sigma}}, s+l)\|\boldsymbol{\Delta}_{\mathcal{S} \cup \mathcal{S}_{1}^{c}}\|_{\mathrm{F}} - \lambda_{\max }(\hat{\boldsymbol{\Sigma}}, l) (\frac{5 \delta^2}{\rho \lambda_K l^{1/2}} + 9 s l^{-1/2} \|\boldsymbol{\Delta}_{\mathcal{S}}\|_{1,1}) \\
    & \geq (\lambda_{\min }(\hat{\boldsymbol{\Sigma}}, s+l) - 9 s l^{-1/2} \lambda_{\max }(\hat{\boldsymbol{\Sigma}}, l)) \|\boldsymbol{\Delta}_{\mathcal{S} \cup \mathcal{S}_{1}^{c}}\|_{\mathrm{F}} - \frac{5 \lambda_{\max }(\hat{\boldsymbol{\Sigma}}, l) \delta^2}{\rho \lambda_K l^{1/2}}. 
\end{aligned}
\end{equation}

By Condition (C6) and Lemma 6 in \cite{Kean2018convex}, let $c_1 = l / s^2$, for constant $C$, we have 
\begin{equation}
\label{b3.12}
    c^{-1} - C (\frac{(c_1 s^2 + s) \log (ed)}{N})^{1/2} \leq \lambda_{\min }(\hat{\boldsymbol{\Sigma}}, s+l) \leq \lambda_{\max }(\hat{\boldsymbol{\Sigma}}, s+l) \leq c + C (\frac{(c_1 s^2 + s) \log (ed)}{N})^{1/2}. 
\end{equation}
with probability being at least $1 - \exp (C^{\prime} (c_1 s^2 + s) \log (ed))$. 

Substituting (\ref{b3.12}) into (\ref{b3.11}) and using the assumption that $N > C_1 s^2 \log d/ \lambda_K^2$, we have that $\lambda_{\min }(\hat{\boldsymbol{\Sigma}}, s+l) - 9 s l^{-1/2} \lambda_{\max }(\hat{\boldsymbol{\Sigma}}, l)$ is lower bounded by some constant $C_2$ when $c_1$ is sufficiently large. 
Also, $\lambda_{\max }(\hat{\boldsymbol{\Sigma}}, l)$ is bounded by some constant $C_3$ similarly. 
Thus we have
\begin{equation}
\label{b3.13}
    \|\hat{\boldsymbol{\Sigma}}^{1 / 2} \boldsymbol{\Delta} \hat{\boldsymbol{\Sigma}}^{1 / 2}\|_{\mathrm{F}} \geq C_2 \|\boldsymbol{\Delta}_{\mathcal{S} \cup \mathcal{S}_{1}^{c}}\|_{\mathrm{F}} - C_3 \frac{\delta^2}{\rho \lambda_K l^{1/2}}. 
\end{equation}

Now, we have obtained the upper and lower bound of $\|\hat{\boldsymbol{\Sigma}}^{1 / 2} \boldsymbol{\Delta} \hat{\boldsymbol{\Sigma}}^{1 / 2}\|_{\mathrm{F}}$. 
Combining (\ref{b3.5}) and (\ref{b3.13}) and using the fact that $\|\boldsymbol{\Delta}_{\mathcal{S}}\|_{1,1} \leq \|\boldsymbol{\Delta}_{\mathcal{S} \cup \mathcal{S}_{1}^{c}}\|_{1,1} \leq s \|\boldsymbol{\Delta}_{\mathcal{S} \cup \mathcal{S}_{1}^{c}}\|_{\mathrm{F}}$, we have
\begin{equation*}
    C_2 \|\boldsymbol{\Delta}_{\mathcal{S} \cup \mathcal{S}_{1}^{c}}\|_{\mathrm{F}} \leq C_3 \frac{\delta^2}{\rho \lambda_K l^{1/2}} + (\frac{4 \delta^{2}}{\lambda_{K}^{2}}+\frac{6 s \rho}{\lambda_{K}}\|\boldsymbol{\Delta}_{\mathcal{S} \cup \mathcal{S}_{1}^{c}}\|_{\mathrm{F}})^{1/2}. 
\end{equation*}
Squaring both sides, we get
\begin{equation*}
    \|\boldsymbol{\Delta}_{\mathcal{S} \cup \mathcal{S}_{1}^{c}}\|_{\mathrm{F}}^2 \leq C_4 \frac{\delta^4}{\rho^2 \lambda_K^2 l} + C_5 \frac{\delta^{2}}{\lambda_{K}^{2}} + C_6 \frac{s \rho}{\lambda_{K}}\|\boldsymbol{\Delta}_{\mathcal{S} \cup \mathcal{S}_{1}^{c}}\|_{\mathrm{F}}. 
\end{equation*}
Thus, we have 
\begin{equation}
\label{b3.14}
    \|\boldsymbol{\Delta}_{\mathcal{S} \cup \mathcal{S}_{1}^{c}}\|_{\mathrm{F}}^2 \leq C_7 (\frac{s^2 \rho^2}{\lambda_{K}^2} + \frac{\delta^4}{\rho^2 \lambda_K^2 l} +  \frac{\delta^{2}}{\lambda_{K}^{2}}). 
\end{equation}

Now, we can solve for the upper bound of $\| \boldsymbol{\Delta} \|_{\mathrm{F}}$. 
By (\ref{b3.9}) and (\ref{b3.14}), we have
\begin{equation}
\label{b3.15}
\begin{aligned}
    \| \boldsymbol{\Delta} \|_{\mathrm{F}} & \leq \|\boldsymbol{\Delta}_{\mathcal{S} \cup \mathcal{S}_{1}^{c}}\|_{\mathrm{F}} + \sum_{j=2}^{J} \|\boldsymbol{\Delta}_{\mathcal{S}_{j}^{c}}\|_{\mathrm{F}} \\
    & \leq \|\boldsymbol{\Delta}_{\mathcal{S} \cup \mathcal{S}_{1}^{c}}\|_{\mathrm{F}} + \frac{5 \delta^2}{\rho \lambda_K l^{1/2}} + 9 s l^{-1/2} \|\boldsymbol{\Delta}_{\mathcal{S} \cup \mathcal{S}_{1}^{c}}\|_{\mathrm{F}} \\
    & \leq C_8 (\frac{s^2 \rho^2}{\lambda_{K}^2} + \frac{\delta^4}{\rho^2 \lambda_K^2 l} +  \frac{\delta^{2}}{\lambda_{K}^{2}})^{1/2} + \frac{5 \delta^2}{\rho \lambda_K l^{1/2}}. 
\end{aligned}
\end{equation}

Recall that $\delta = \|\tilde{\boldsymbol{\Lambda}} - \boldsymbol{\Lambda}\|_{\mathrm{F}} \leq K^{1/2} \|\tilde{\boldsymbol{\Lambda}} - \boldsymbol{\Lambda}\|_2$. 
By Lemma \ref{lem1} and Lemma 2 in \cite{Kean2018convex}, we have $\|\tilde{\boldsymbol{\Lambda}} - \boldsymbol{\Lambda}\|_2 \leq C (s / N)^{1/2}$ with probability being greater than $1 - \exp(- C^{\prime} s)$. 
Thus $\delta \leq C (\frac{K s}{N})^{1/2}$. 
Under Condition (C5) and the assumption that $\rho \geq C'' (\log d/ N)^{1/2}$ for some constant $C''$, we have $\delta \leq C_9 \rho l^{1/4}$. 
Substituting this into (\ref{b3.15}), we have
\begin{equation}
    \| \boldsymbol{\Delta} \|_{\mathrm{F}} \leq C_8 (\frac{s^2 \rho^2}{\lambda_{K}^2} + C_9^4 \frac{\rho^2 }{\lambda_K^2} + C_9^2 \frac{s \rho^{2}}{\lambda_{K}^{2}})^{1/2} + C_9^2 \frac{5 \rho}{\lambda_K} \leq C_{10} \frac{s \rho}{\lambda_K}. 
\end{equation}
Thus, we obtain the upper bound for $\| \hat{\boldsymbol{\Pi}} - \boldsymbol{\Pi} \|_{\mathrm{F}}$. 
Combined with (\ref{b3.0}), for some constant $C$, we have
\begin{equation}
    \| \hat{\boldsymbol{\Pi}} - \boldsymbol{\Pi} \|_{\mathrm{F}} \leq C \frac{s \rho}{\lambda_K}, 
\end{equation}
with probability being at least $1 - \exp(-C^{\prime} s)$. 

From Corollary 3.2 in \cite{vu2013fantope}, for the population central subspace $\cV$ and the estimated subspace $\hat{\cV}$, we have 
\begin{equation}
    D(\cV, \hat{\cV}) \leq C \frac{s \rho}{\lambda_K}, 
\end{equation}
with probability being at least $1 - \exp(-C^{\prime} s)$. 
Combining Corollary \ref{cor1} with Lemma 7 in \cite{Kean2018convex}, we can choose $\rho \leq C_1^{\prime} (\log d / N)^{1/2} + C_2^{\prime} (\log d)^{1/2} / N^{1 - \eta}$, then $\|\hat{\mathbf{T}} - \tilde{\mathbf{T}} \|_{\max} \leq \frac{\rho}{2}$ holds with probability at least $1 - \exp(-C_3^{\prime} \log d)$, which completes the proof. 
\end{proof}

\section{Dimension determination and other simulation results}
\label{bic}

\subsection{Proof of Corollary \ref{cor:bic}}

We first give the proof of Corollary \ref{cor:bic}. 

\begin{proof}
Theorem 6 in \cite{zhu2010cumulative} states that $\hat{K}_i$ converges to $K$ in probability, $\forall \gamma > 0$, there exists $N_i > 0$ that for $n_i > N_i$, $$\mathrm{Pr} (\hat{K}_i \neq K) < \gamma. $$
Suppose there are $k$ clients making correct decision of $K$ but $\hat{K} \neq K$, then there must exist a $K^{\prime}$ that at least $k+1$ clients choosing $K^{\prime}$. 
This means if $\hat{K} \neq K$, there are at least $\lfloor \frac{m}{2} \rfloor$ clients making the wrong decision. 
Thus, let $\hat{K}_{1}^{\prime}, \dots, \hat{K}_{m}^{\prime}$ be a permutation of $\hat{K}_{1}, \dots, \hat{K}_{m}$, we have 
\begin{align*}
    \mathrm{Pr} (\hat{K} \neq K) & \leq \sum_{i=0}^{\lfloor \frac{m}{2} \rfloor} \binom{m}{i} \mathrm{Pr}(\hat{K}_{1}^{\prime} \neq K) \times \cdots \times \mathrm{Pr}(\hat{K}_{m-i}^{\prime} \neq K) \\
    & \times \mathrm{Pr}(\hat{K}_{m-i+1}^{\prime} = K) \times \cdots \times \mathrm{Pr}(\hat{K}_{m}^{\prime} = K) \\
    & \leq \sum_{i=0}^{\lfloor \frac{m}{2} \rfloor} \binom{m}{i} \gamma^{m-i} 1^{i} \\
    & \leq \sum_{i=0}^{\lfloor \frac{m}{2} \rfloor} \binom{m}{i} \gamma^{\lfloor \frac{m}{2} \rfloor} \\
    & \leq (2 \sqrt{\gamma})^{m}.
\end{align*}
\end{proof}

\subsection{Simulation on non-Gaussian covariates}

We conduct a simulation to show the performance of our proposal on non-Gaussian covariates. 
We generate non-Gaussian covariates in the following way. 
Let $\bx^{(i)} = \bx_{1}^{(i)} + \bx_{2}^{(i)}$, where $\bx_{1}^{(i)}$ is generated same as before and the elements of $\bx_{2}^{(i)}$ are sampled independently from Bernoulli distribution. 
Since $\bx_{1}^{(i)}$ and $\bx_{2}^{(i)}$ are independently sampled from sub-Gaussian distributions, $\bx^{(i)}$ is also a sub-Gaussian random vector, but not Gaussian. 

\begin{table}[ht]
\begin{center}
\begin{minipage}{\textwidth}
\caption{True and false positive rates, and subspace distances with $\alpha = 5$, $N = 2000$, $m = 10$, $d = 100$. 
All entries are averaged across 200 runs. 
The standard deviations are in the brackets. 
}
\label{tabNG}
\begin{tabular}{@{}cccccccc@{}}
\toprule
            &      & Setting 1 & Setting 2 & Setting 3 & Setting 4 & Setting 5 & Setting 6 \\ \midrule
FedSSIR  	& TPR  & 1.000 & 1.000 & 0.963 & 0.851 & 1.000 & 0.950 \\ 
            &      & (0) & (0) & (0.117) & (0.210) & (0) & (0.133) \\ 
			& FPR  & 0.000 & 0.002 & 0.005 & 0.007 & 0.000 & 0.011 \\ 
            &      & (0.001) & (0.005) & (0.011) & (0.012) & (0.002) & (0.013) \\ 
			& Dist & 0.050 & 0.075 & 0.462 & 0.683 & 0.059 & 0.459 \\ 
            &      & (0.030) & (0.035) & (0.277) & (0.348) & (0.029) & (0.246) \\ 
SSIR  		& TPR  & 1.000 & 1.000 & 0.764 & 0.571 & 1.000 & 0.989 \\ 
            &      & (0) & (0) & (0.250) & (0.386) & (0) & (0.064) \\ 
			& FPR  & 0.928 & 0.918 & 0.517 & 0.280 & 0.199 & 0.032 \\ 
            &      & (0.092) & (0.105) & (0.269) & (0.265) & (0.330) & (0.147) \\ 
			& Dist & 1.819 & 1.544 & 1.802 & 1.620 & 0.877 & 0.488 \\ 
            &      & (0.134) & (0.251) & (0.166) & (0.152) & (0.538) & (0.333) \\
LassoSIR  	& TPR  & 1.000 & 1.000 & 0.969 & 0.997 & 1.000 & 0.999 \\ 
            &      & (0) & (0) & (0.090) & (0.032) & (0) & (0.014) \\ 
			& FPR  & 0.327 & 0.270 & 0.682 & 0.991 & 0.395 & 0.996 \\ 
            &      & (0.114) & (0.104) & (0.350) & (0.074) & (0.406) & (0.057) \\ 
			& Dist & 0.136 & 0.141 & 1.777 & 1.776 & 1.176 & 1.328 \\ 
            &      & (0.092) & (0.063) & (0.590) & (0.523) & (0.718) & (0.367) \\ 
\botrule
\end{tabular}
\end{minipage}
\end{center}
\end{table}
We keep $N = 2000$, $d = 100$ and split the data onto $10$ clients using the following heterogeneous set up. 
We generate a random vector $\boldsymbol{\omega} \sim \mathrm{Dirichlet}_m(5)$ and $(n_1, \dots, n_m) \sim \mathrm{Multinomial}(N, \boldsymbol{\omega})$ with $m = 10$. 
The number of dimension reduction directions $K$ is chosen by the BIC-type criterion. 
The penalization parameter $\rho$ is selected by the hold-out validation outlined in section \ref{sec2.3}. 
We also include the results of SSIR and LassoSIR. 
The results are shown in Table \ref{tabNG}. 
We can see that our method still performs well on non-Gaussian covariates. 

\subsection{Dimension determination results}

Here, we also report the proportion of correct decisions about the dimension of the central subspace in our simulations in Table \ref{tab:bic}. 
We can see that BIC type criterion has a satisfactory performance in most cases. 
As the sample size grows, the performance becomes better, which is consistent with our conclusion that $\hat{K}$ converges to $K$ in probability. 
Our proposed BIC has similar performance under different $\boldsymbol{\omega}$ settings, which reflects that our method performs well for unbalanced datasets. 
Compared with the other two methods, our proposed BIC performs more robustly under the non-Gaussian setting. 

\begin{table}[ht]
\begin{minipage}{\textwidth}
\caption{The empirical probabilities of correctly estimating dimension in different simulations. }
\label{tab:bic}
\begin{center}
\begin{tabular}{@{}cccccccc@{}}
\toprule
 & & Setting 1 & Setting 2 & Setting 3 & Setting 4 & Setting 5 & Setting 6 \\ \midrule
 n = 100 & BIC & 1.000 & 0.950 & 0.785 & 0.855 & 1.000 & 0.805 \\ 
 & SSIR & 0.020 & 0.350 & 0.575 & 0.065 & 0.430 & 0.530 \\
 & LassoSIR & 1.000 & 1.000 & 0.030 & 0.055 & 1.000 & 0.525 \\
 n = 200 & BIC & 1.000 & 0.995 & 0.825 & 0.865 & 1.000 & 0.875 \\
 & SSIR & 0.015 & 0.290 & 0.450 & 0.065 & 0.410 & 0.615 \\
 & LassoSIR & 0.995 & 1.000 & 0.035 & 0.100 & 1.000 & 0.995 \\ \midrule
 $\alpha=1$ & BIC & 1.000 & 1.000 & 0.960 & 0.985 & 1.000 & 0.980 \\
 $\alpha=2$ & BIC & 1.000 & 1.000 & 0.985 & 0.995 & 1.000 & 0.995 \\ 
 $\alpha=5$ & BIC & 1.000 & 1.000 & 0.995 & 1.000 & 1.000 & 1.000 \\ \midrule
 non-Gaussian & BIC & 1.000 & 1.000 & 0.840 & 0.760 & 1.000 & 0.870 \\ 
 & SSIR & 0.015 & 0.385 & 0.645 & 0.045 & 0.365 & 0.790 \\ 
 & LassoSIR & 0.995 & 1.000 & 0.025 & 0.010 & 0.700 & 0.010 \\ 
\botrule
\end{tabular}
\end{center}
\end{minipage}
\end{table}

\section{Auxiliary definitions}
\label{defs}

Here are some definitions used in the text. 
Let $y_{(1)}, \dots, y_{(n)}$ be the order statistics of $y_1, \dots, y_n$. 
Define the inverse regression function $m(y) = \bE[\bx \vert y]$ and let $m(y_{(k)}) = \bE[\bx \vert y_{(k)}]$. 
Total variation for a vector-valued function $m(y)$ under the $L_{\infty}$ norm is defined as below.  
\begin{definition}
\label{def1}
Let $\cU(B)$ be the collection of all $n$-point partitions $-B \leq y_{(1)} \leq \dots \leq y_{(n)} \leq B$ on the interval $[-B, B]$, where $B > 0$ and $n \geq 1$. 
A vector-valued function $m(y)$ is said to have a total variation defined as $\sup_{\cU(B)} \sum_{k=2}^{n} \| m(y_{(k)}) - m(y_{(k-1)})\|_{\infty}$. 
\end{definition}

Meinshausen and Yu \cite{Meinshausen2009lassohigh} gave a definition on the $s$-sparse eigenvalue of a matrix $\mathbf{M}$. 
\begin{definition}
\label{def2}
The $s$-sparse minimal and maximal eigenvalues of $\mathbf{M}$ are $$\lambda_{\min} (\mathbf{M}, s) = \min_{v:\|v\|_0 \leq s} \frac{v^{\mathrm{T}} \mathbf{M} v}{v^{\mathrm{T}} v}, \quad \quad \lambda_{\max} (\mathbf{M}, s) = \max_{v:\|v\|_0 \leq s} \frac{v^{\mathrm{T}} \mathbf{M} v}{v^{\mathrm{T}} v}. $$
\end{definition}

There are two consistent estimators for $\mathbf{Q}_i$. 
Hsing and Carroll \cite{hsing1992asym} gave a two slice estimation $\hat{\mathbf{Q}}_i$. 
Let $y^{(i)}_{(1)}, \dots, y^{(i)}_{(n_i)}$ be the order statistics of responses. 
$\bx^{(i)}_{(k)^*}$ is denoted as the value of $\bx$ associated with $y^{(i)}_{(k)}$ and termed the concomitant of $y^{(i)}_{(k)}$. 
Then the estimator is 
\begin{equation}
    \label{eq2.6}
    \hat{\mathbf{Q}}_i = \frac{1}{n_i} \sum_{k=1}^{\lfloor n_i/2 \rfloor} \{\bx^{(i)}_{(2k)^*} - \bx^{(i)}_{(2k-1)^*}\}\{\bx^{(i)}_{(2k)^*} - \bx^{(i)}_{(2k-1)^*}\}^{\mathrm{T}}, 
\end{equation}
where $\lfloor \cdot \rfloor$ is the floor function that returns the greatest integer less than or equal to the input. 

Zhu and Ng \cite{zhu1995asymptotics} gave another estimator for $\mathbf{Q}_i$. 
First, partition the dataset $S_i$ into $H$ slices with respect to the order statistics of $y^{(i)}$ and let $\Xi_1, \dots, \Xi_H$ be the indice sets for each slice. 
Then, the estimator $\tilde{\mathbf{Q}}_i$ is obtained by computing the average of covariance matrices on each slice: 
\begin{equation}
    \label{eq2.6a}
    \tilde{\mathbf{Q}}_i = \frac{1}{H} \sum_{h=1}^{H} \{ \frac{1}{n_h} \sum_{l \in \Xi_h} (\bx^{(i)}_l - \bar{\bx}^{(i)}_{\Xi_h}) (\bx^{(i)}_l - \bar{\bx}^{(i)}_{\Xi_h})^{\mathrm{T}} \}. 
\end{equation}
Therefore, a trivial estimator for $\mathbf{T}_i$ is $\hat{\mathbf{T}}_i = \hat{\boldsymbol{\Sigma}}_i - \hat{\mathbf{Q}}_i$. 

Similar to \ref{sec2.1}, we derive the population and empirical average conditional covariance matrices. 
Let $\boldsymbol{\epsilon} = \bx - \bE[\bx \vert y]$, the population average conditional covariance matrix $\bar{\mathbf{Q}}$ can be represented as $\bar{\mathbf{Q}} = \boldsymbol{\Sigma} - \bar{\mathbf{T}} = \sum_{i=1}^{m} \omega_i (\boldsymbol{\Sigma}_i - \mathbf{T}_i) = \sum_{i=1}^{m} \omega_i \mathbf{Q}_i$. 
Thus, the empirical average conditional covariance matrix $\hat{\mathbf{Q}} = \sum_{i=1}^{m} \hat{\omega}_i \hat{\mathbf{Q}}_i = \sum_{i=1}^{m} \frac{n_i}{N} \hat{\mathbf{Q}}_i$. 




\end{appendices}


\bibliography{sn-bibliography}


\end{document}